\documentclass{article} 
\usepackage{iclr2027_conference,times}

\usepackage{amsmath,amsfonts,bm}

\def\eqref#1{equation~\ref{#1}}

\def\1{\bm{1}}

\DeclareMathAlphabet{\mathsfit}{\encodingdefault}{\sfdefault}{m}{sl}
\SetMathAlphabet{\mathsfit}{bold}{\encodingdefault}{\sfdefault}{bx}{n}

\usepackage{hyperref}
\usepackage{url}

\usepackage{amsmath}
\usepackage{amssymb}
\usepackage{algorithm}
\usepackage{algpseudocode}
\usepackage{booktabs}
\usepackage{graphicx}
\usepackage{tabularx}
\usepackage{multirow}
\usepackage[table]{xcolor}

\newif\ifshowmarks
\showmarkstrue      

\title{Less Data, Better Timing: Student--Curriculum Coupling for VLM On-Policy Distillation in Temporal Video Grounding}

\author{%
  \makebox[\dimexpr\textwidth-2\tabcolsep\relax][c]{%
    \begin{tabular}[t]{@{}c@{}}
      {\normalfont\normalsize\bfseries
        Jiacheng~Qiu, Yunsoo~Kim, Ruichen~Xu,
        Jian~Luo, Petar~M.~Djurić, Sima~Mofakham}\\[4pt]
      {\normalfont\small
        State University of New York at Stony Brook}
    \end{tabular}%
  }%
}

\iclrfinalcopy 

\begin{document}
\maketitle

\pagestyle{fancy}
\fancyhead{}
\fancyhead[L]{Preprint}
\thispagestyle{fancy}


\vspace{-1em}
\begin{abstract}

On-policy distillation (OPD) provides dense supervision directly on student-generated trajectories, making it an effective post-training strategy for vision-language models in temporal video grounding (TVG). However, existing pipelines typically construct the training curriculum from a fixed teacher and the initial student state, implicitly assuming that selected examples retain positive supervision value throughout optimization. We show that supervision trustworthiness and supervision necessity are distinct yet coupled: the former concerns target credibility, while the latter varies with the student's current task competence; together, they shape supervision value. Building on this coupled view, we introduce \textbf{Student--Curriculum Coupling (SCC)}, a closed-loop framework in which a compact Anchor--Frontier curriculum defines the candidate supervision space and the evolving student dynamically determines its active subset. Supervision can therefore be activated, suspended, or reactivated as competence changes, concentrating teacher computation and optimization on current task-level deficits. Across three TVG benchmarks, SCC achieves a 5.1\% relative improvement in mean recall over Video-OPD on its original curriculum, while using 60.0\% fewer training examples and reducing training time by 50.4\%. Ablations support the complementary roles of capability-structured curriculum design and student-dependent supervision in achieving these gains. Together, these results establish SCC as a data- and compute-efficient framework for TVG post-training, delivering stronger temporal grounding by aligning trustworthy supervision with the student's evolving learning needs.

\vspace{-0.5em}
\end{abstract}

\section{Introduction}
\vspace{-0.5em}

Temporal video grounding (TVG) localizes the segment in an untrimmed video that corresponds to a natural-language query~\citep{gao2017tall}. As vision-language models (VLMs) have become increasingly capable of jointly reasoning over video and language, recent work has cast TVG as generative temporal prediction, with target intervals produced directly as timestamps~\citep{huang2024vtimellm,wang2025grounded}. However, precise localization, particularly at event boundaries, remains challenging, motivating task-specific post-training~\citep{wang2025timer1}. Among these approaches, on-policy distillation (OPD) trains the student on its own generated trajectories using dense token-level teacher feedback~\citep{lu2025onpolicydistillation,agarwal2024onpolicy,li2026rethinking}. Video-OPD applies this paradigm to TVG~\citep{li2026video}. Beyond the distillation objective itself, the effectiveness of OPD for TVG also depends on how teacher supervision is allocated across training examples, highlighting the importance of curriculum design and adaptive data selection~\citep{bengio2009curriculum,chen2023skillit}.

In OPD for TVG, curriculum construction can use teacher reliability and the performance gap between a fixed teacher and the initial student to select informative examples~\citep{li2026video}. Selected examples typically receive supervision whenever presented, carrying an implicit \emph{persistent-value assumption} that their supervision remains useful throughout optimization. As student competence evolves, however, initial suitability alone cannot determine whether supervision remains necessary. We therefore distinguish \emph{supervision trustworthiness}, which concerns whether the teacher provides a credible target for an example, from \emph{supervision necessity}, which concerns whether the current student still exhibits a task-level deficit on that example. Together, these factors motivate a coupled view in which curriculum composition and supervision realization are jointly designed to align trustworthy supervision with the student's changing learning needs.

Building on this coupled view, we introduce \textbf{Student--Curriculum Coupling (SCC)}, a closed-loop OPD framework that jointly designs curriculum composition and supervision realization. SCC constructs a compact Anchor--Frontier (AF) curriculum around the initial student's capability profile, preserving trustworthy supervision opportunities for capability acquisition and stabilization. These opportunities contribute to optimization only when the current student exhibits a task-level deficit, so that candidate-space composition and current student competence jointly determine active supervision. OPD updates in turn reshape the student's supervision needs, which guide subsequent curriculum realization and close the feedback loop.

We evaluate SCC on Charades-TimeLens~\citep{zhang2025timelens}, ActivityNet-TimeLens~\citep{zhang2025timelens}, and QVHighlights-TimeLens~\citep{zhang2025timelens}. Compared with Video-OPD on its Teacher-Validated Disagreement Focusing (TVDF) curriculum~\citep{li2026video}, SCC achieves a 5.1\% relative improvement in mean recall across the three benchmarks and intersection-over-union (IoU) thresholds of $0.3$, $0.5$, and $0.7$. These gains are obtained with 60.0\% fewer training examples and 50.4\% shorter training time.

\noindent Our contributions are threefold:
\par\vspace{1pt}
\begingroup
\setlength{\parskip}{0pt}
\newcommand{\contribitem}[1]{%
    \par\noindent
    \hangindent=1.4em
    \hangafter=1
    \makebox[1.4em][l]{\textbullet}#1\par
}
\contribitem{We identify the implicit persistent-value assumption in TVG OPD and develop a coupled view that links supervision trustworthiness with the student's evolving supervision necessity.}
\contribitem{We introduce SCC, a closed-loop OPD framework that couples a compact AF candidate space with student-dependent supervision realization.}
\contribitem{We demonstrate consistent gains in TVG accuracy and training efficiency over Video-OPD on TVDF. Ablations isolate the roles of candidate-space composition and student-dependent realization, while training-dynamics analysis tracks supervision as student competence evolves.}
\endgroup

\vspace{-0.5em}
\section{Problem Formulation and Motivation}
\label{sec:motivation}

\vspace{-0.5em}

\begin{figure}[!t]
    \centering
    \setlength{\abovecaptionskip}{4pt}
    \includegraphics[width=\linewidth]{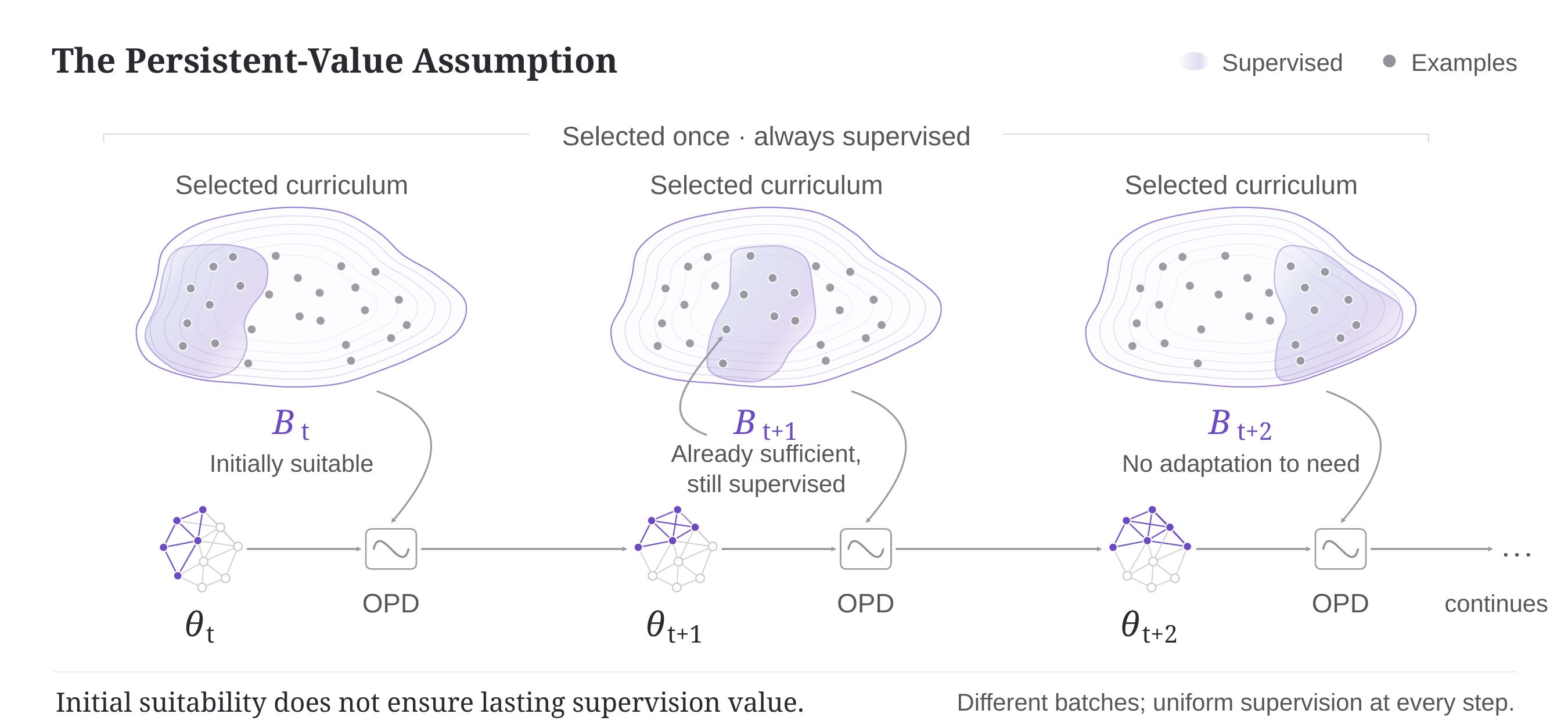}
    \vspace{-1em}
    \caption{\textbf{The persistent-value assumption.} All examples in each scheduled batch (shaded) receive OPD supervision as the student evolves, including those that already satisfy the task criterion.
    }
    \label{fig:motivation}
    \vspace{-0.8em}
\end{figure}

Our analysis focuses on OPD in VLM post-training for TVG. Section~\ref{sec:static_supervision} identifies the persistent-value assumption illustrated in Figure~\ref{fig:motivation}; Section~\ref{sec:nonstationary_value} formalizes the non-stationary marginal value of supervision; and Section~\ref{sec:trustworthiness_necessity} distinguishes supervision trustworthiness from necessity.

\vspace{-0.5em}
\subsection{The Assumption of Persistent Supervision Value}
\label{sec:static_supervision}

For a TVG example $x=(v,q)$, where $v$ is an untrimmed video and $q$ is a natural-language query, let $a^\star(x)$ denote the ground-truth temporal interval and $a_M(x)$ the interval predicted by model $M$ under a fixed decoding protocol. A common curriculum-construction strategy evaluates each candidate using a fixed teacher and the initial student, and selects examples according to their TVG performance:
\begin{equation}
q_M(x)
=
\operatorname{tIoU}\!\left(a_M(x),a^\star(x)\right),
\qquad
\mathcal{D}_{\mathrm{cand}}
=
\left\{ x\in\mathcal{P} : c\!\left(q_T(x),q_{S,0}(x)\right)=1 \right\}.
\label{eq:static_curriculum}
\end{equation}
Here, $\operatorname{tIoU}(\cdot,\cdot)$ denotes temporal intersection over union, which measures the overlap between two temporal intervals. Accordingly, $q_M(x)\in[0,1]$ is the TVG score of model $M$ on example $x$. $q_T(x)$ and $q_{S,0}(x)$ denote the TVG scores of the fixed teacher and the student before optimization, respectively. $\mathcal{P}$ is the candidate data pool, $c:[0,1]^2\rightarrow\{0,1\}$ is the selection rule, and $\mathcal{D}_{\mathrm{cand}}\subseteq\mathcal{P}$ is the resulting candidate curriculum.

This construction selects examples with a credible teacher signal and a targeted level of difficulty for the initial student. However, $q_{S,0}(x)$ reflects the initial student's performance on $x$, not a fixed property of the example. Once $x$ is included in $\mathcal{D}_{\mathrm{cand}}$, its curriculum membership remains unchanged even as optimization alters the student's competence and the potential learning value of $x$. Continuing to treat $x$ as supervision-bearing whenever it is scheduled therefore introduces an implicit assumption of persistent supervision value.

\vspace{-0.4em}
\subsection{The Non-Stationary Marginal Value of Supervision}
\label{sec:nonstationary_value}

To formalize how supervision value changes with the student, let $\theta_t$ denote the student parameters immediately before training step $t$, and let $J_{\mathrm{TVG}}(\theta)$ denote the expected TVG generalization performance of the student parameterized by $\theta$. For an example $x$ encountered at step $t$, let $\theta_{t+1}^{+x}$ denote the student parameters after an update that includes the OPD loss associated with $x$. Similarly, let $\theta_{t+1}^{-x}$ denote the parameters after the corresponding update without the OPD loss associated with $x$. All other examples and optimization conditions are held fixed. We define the marginal supervision value of $x$ as:
\begin{equation}
\Delta_t(x)
=
\mathbb{E}\!\left[
J_{\mathrm{TVG}}\!\left(\theta_{t+1}^{+x}\right)
-
J_{\mathrm{TVG}}\!\left(\theta_{t+1}^{-x}\right)
\mid
\theta_t,x
\right].
\label{eq:marginal_supervision_value}
\end{equation}
The expectation is taken over training stochasticity, with the same realization used for the updates with and without OPD supervision on $x$. This quantity characterizes the effect of supervising $x$ and is not assumed to be directly observable during training.

Even when the video, query, annotation, and teacher remain fixed, $\Delta_t(x)$ can vary because the student-generated trajectory, the resulting OPD signal, and the model's response to that signal all depend on $\theta_t$. Supervision may support capability acquisition before the student can localize the queried event, lose value once the student produces a satisfactory interval, and become useful again if later updates degrade that capability. The marginal supervision value is therefore non-stationary and reflects both the student's training history and its current interaction with the example.


\vspace{-0.4em}
\subsection{Supervision Trustworthiness and Necessity}
\label{sec:trustworthiness_necessity}

We characterize the state dependence of supervision value through \emph{supervision trustworthiness} $\mathcal{T}(x)$ and \emph{supervision necessity} $\mathcal{N}_t(x)$. The former indicates whether the teacher provides a credible target for example $x$, while the latter indicates whether the current student still exhibits a task-level deficit:
\begin{equation}
\mathcal{T}(x)
=
\mathbf{1}\!\left\{q_T(x)\geq\tau_T\right\},
\qquad
\mathcal{N}_t(x)
=
\mathbf{1}\!\left\{q_{S,t}(x)<\tau_S\right\}.
\label{eq:trustworthiness_necessity}
\end{equation}
$\mathbf{1}\{\cdot\}$ denotes the indicator function. $q_{S,t}(x)$ is the student's TVG score on $x$ at step $t$, and $\tau_T$ and $\tau_S$ are thresholds for supervision trustworthiness and student competence, respectively. $\mathcal{T}(x)$ captures the fixed teacher--example relationship, whereas $\mathcal{N}_t(x)$ captures the evolving student--example relationship. The necessity indicator is a task-level proxy and does not directly observe $\Delta_t(x)$.



The two factors are distinct but coupled in supervision allocation. When both hold, a trustworthy target is available for an unresolved student outcome. Trustworthiness without necessity may render further supervision redundant (Figure~\ref{fig:motivation}), while necessity without trustworthiness provides no credible distillation target. Curriculum composition therefore determines the availability of trustworthy supervision, while current student competence determines its necessity during optimization.

\vspace{-0.5em}

\section{Student--Curriculum Coupling}
\label{sec:method}
\vspace{-0.5em}

Building on the coupled view developed in
Section~\ref{sec:motivation}, we introduce Student--Curriculum Coupling,
a closed-loop OPD framework for TVG. Section~\ref{sec:opd} reviews the OPD objective, and Section~\ref{sec:af_candidate_space} introduces the Anchor--Frontier candidate supervision space. Section~\ref{sec:student_dependent_realization} defines student-dependent realization, while Section~\ref{sec:closed_loop_optimization} presents the coupled objective and efficient execution. Figure~\ref{fig:method_overview} illustrates the framework. Additional theoretical details are provided in Appendix~\ref{app:all}.

\begin{figure}[!t]
    \centering
    \setlength{\abovecaptionskip}{4pt}
    \includegraphics[width=\linewidth]{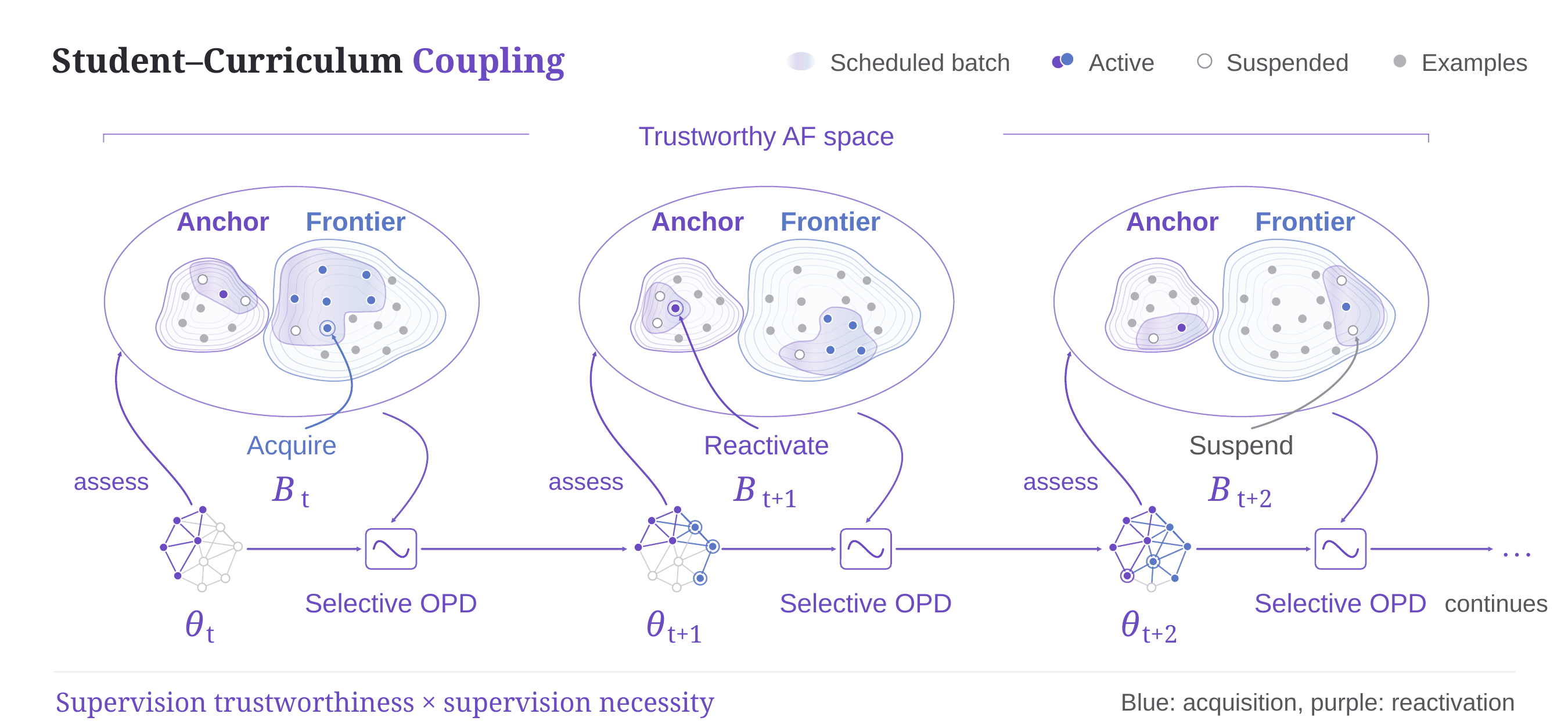}
    \vspace{-1em}
    \caption{\textbf{Student--Curriculum Coupling.}
    Shaded regions denote successive scheduled batches within the AF space. Current-student assessment determines active and suspended supervision, supporting capability acquisition and Anchor reactivation. Selective OPD updates the student, whose new state guides supervision in subsequent batches, closing the loop.}
    \label{fig:method_overview}
    \vspace{-0.8em}
\end{figure}

\vspace{-0.5em}
\subsection{On-Policy Distillation}
\label{sec:opd}


On-policy distillation optimizes a student VLM $\pi_\theta$ using token-level supervision from a teacher VLM $\pi_T$ whose parameters remain fixed. The student generates trajectories from its current policy, and the teacher supplies target next-token distributions at the corresponding prefixes~\citep{agarwal2024onpolicy,li2026video}. Given a TVG input $x=(v,q)$, the rollout policy $\pi_{\theta_t}$
at training step $t$ generates an autoregressive trajectory of length $L_t$:
\[
y_t=(y_{t,1},\ldots,y_{t,L_t})
\sim\pi_{\theta_t}(\cdot\mid x).
\]
At position $k$, let $h_{t,k}=(x,y_{t,<k})$ denote the current prefix, and define the student and teacher next-token distributions as
$p_{\theta,k}(\cdot)=\pi_\theta(\cdot\mid h_{t,k})$ and
$p_{T,k}(\cdot)=\pi_T(\cdot\mid h_{t,k})$, respectively.
The frozen teacher scores the student-generated trajectory without a separate teacher rollout.


Using reverse Kullback--Leibler (KL) divergence for token-level alignment~\citep{gu2024minillm,li2026rethinking}, we define the per-example objective under the fixed rollout distribution as:
\begin{equation}
\mathcal{L}_{\mathrm{OPD},t}(\theta;x)
=
\mathbb{E}_{y_t\sim\pi_{\theta_t}(\cdot\mid x)}
\left[
\sum_{k=1}^{L_t}
D_{\mathrm{KL}}
\!\left(
p_{\theta,k}
\,\middle\|\,
p_{T,k}
\right)
\right].
\label{eq:opd_objective}
\end{equation}
Here, $\theta_t$ denotes the fixed rollout-policy parameters, while $\theta$ is the optimization variable initialized from $\theta_t$. The rollout distribution remains fixed during differentiation, and gradients propagate through the student next-token probabilities.

Following Video-OPD~\citep{li2026video}, we optimize an importance-weighted sampled-token surrogate
$\ell_{\mathrm{OPD}}(x,y_t;\theta,\theta_t,\pi_T)$, with teacher-derived signals held fixed during the update. At $\theta=\theta_t$, its gradient matches that of Equation~\ref{eq:opd_objective} in expectation over student rollouts. Appendix~\ref{app:opd_estimator} provides the surrogate and derives this
local gradient identity.

\vspace{-0.4em}
\subsection{Anchor--Frontier Candidate Space}
\label{sec:af_candidate_space}

To support capability acquisition and stabilization within a compact supervision space, we introduce the Anchor--Frontier (AF) design. It combines examples already supported by the initial student with examples that offer substantial learning headroom. Both groups contain only examples with trustworthy supervision, satisfying $\mathcal{T}(x)=1$ as defined in Equation~\ref{eq:trustworthiness_necessity}.

\begin{subequations}
\label{eq:anchor_frontier_space}
Anchors retain examples on which the initial student already performs well. They preserve access to corrective supervision if the student's performance on these examples later declines. From the candidate pool $\mathcal{P}$, we select $N_A$ such examples:
\begin{equation}
\mathcal{A}
=
\operatorname{Select}_{N_A}
\left(
\left\{
x\in\mathcal{P}
:
\mathcal{T}(x)=1,\;
q_{S,0}(x)\geq\tau_A
\right\}
\right).
\label{eq:anchor_set}
\end{equation}
Frontiers complement this coverage with examples on which the initial student has low competence, providing opportunities for capability acquisition. We select $N_F$ such examples:
\begin{equation}
\mathcal{F}
=
\operatorname{Select}_{N_F}
\left(
\left\{
x\in\mathcal{P}
:
\mathcal{T}(x)=1,\;
q_{S,0}(x)<\tau_F
\right\}
\right).
\label{eq:frontier_set}
\end{equation}
\end{subequations}
Here, $N_A$ and $N_F$ are the selection budgets, while $\tau_A$ and $\tau_F$ define the high- and low-competence regions. The operator $\operatorname{Select}_{N}(\cdot)$ returns a subset of $N$ examples. The thresholds satisfy $\tau_F<\tau_A$, leaving the intermediate competence range outside the candidate space. The candidate space, $\mathcal{D}_{\mathrm{AF}}=\mathcal{A}\cup\mathcal{F}$, remains fixed during training. Section~\ref{sec:student_dependent_realization} describes how its supervision is realized as the student evolves.

\vspace{-0.4em}
\subsection{Student-Dependent Realization}
\label{sec:student_dependent_realization}

We organize the AF candidate space into a fixed training schedule $\mathcal{S}=(\mathcal{B}_0,\ldots,\mathcal{B}_{K-1})$, where $\mathcal{B}_t\subseteq\mathcal{D}_{\mathrm{AF}}$ contains the examples assigned to step $t$. Here, $K$ counts scheduled training steps, including those that produce no optimizer update. Scheduled batch sizes may vary across steps. The schedule determines when each example is assessed, while the current student determines whether it receives OPD supervision.

At step $t$, the student generates one trajectory $y_{t,i}\sim\pi_{\theta_t}(\cdot\mid x_i)$ for each $x_i\in\mathcal{B}_t$. We assess the current prediction using:
\begin{equation}
q_{S,t}(x_i;y_{t,i})
=
\operatorname{tIoU}
\!\left(a(y_{t,i}),a^\star(x_i)\right),
\label{eq:current_student_assessment}
\end{equation}
where $a(y_{t,i})$ is the temporal interval decoded from the trajectory. Applying the supervision-necessity criterion from Section~\ref{sec:trustworthiness_necessity} yields the effective batch:
\begin{equation}
\mathcal{B}^{\mathrm{eff}}_t
=
\left\{
(x_i,y_{t,i})
:
x_i\in\mathcal{B}_t,\;
q_{S,t}(x_i;y_{t,i})<\tau_S
\right\}.
\label{eq:curriculum_realization}
\end{equation}
Each retained trajectory is reused for teacher scoring and OPD optimization.

This rule supports capability acquisition, supervision suspension, and Anchor reactivation. A Frontier whose current score is below the criterion receives supervision for capability acquisition. For any candidate that meets the criterion, OPD supervision is suspended. For an Anchor with $q_{S,0}(x_i)\geq\tau_S$ at curriculum construction, a current score $q_{S,t}(x_i;y_{t,i})<\tau_S$ reactivates its supervision to support capability stabilization. Reactivation is thus defined relative to the initial competence assessment. 

\vspace{-0.4em}
\subsection{Closed-Loop OPD Optimization}
\label{sec:closed_loop_optimization}

For each nonempty scheduled batch $\mathcal{B}_t$, we aggregate the per-example OPD surrogate from Section~\ref{sec:opd} over the effective batch:
\begin{equation}
\mathcal{L}^{\mathrm{SCC}}_t(\theta)
=
\frac{1}{n_t}
\sum_{(x_i,y_{t,i})\in\mathcal{B}^{\mathrm{eff}}_t}
\ell_{\mathrm{OPD}}
\!\left(x_i,y_{t,i};\theta,\theta_t,\pi_T\right).
\label{eq:coupled_opd_objective}
\end{equation}
Here, $\theta$ is initialized from $\theta_t$ for the current update. The normalization denominator $n_t$ is preset for each step, independently of the number of assigned or retained examples. Appendix~\ref{app:normalization_execution} derives the equivalent masked formulation and explains the effect of this normalization. The rollout policy and effective-batch membership remain fixed during optimization, with no gradients through sampling, temporal decoding, or competence assessment.


Under this objective, every scheduled example requires a student rollout for assessment, but teacher scoring, OPD loss evaluation, and backpropagation are restricted to $\mathcal{B}^{\mathrm{eff}}_t$. The compact AF space limits rollout workload, while student-dependent realization reduces the subsequent supervision cost. If $\mathcal{B}^{\mathrm{eff}}_t=\varnothing$, the optimizer and weight-decay updates are skipped, leaving the student parameters and optimizer state unchanged as the schedule advances.

Let $\mathcal{G}$ denote the sampling, assessment, and selection procedure in Section~\ref{sec:student_dependent_realization}, so that
$\mathcal{B}^{\mathrm{eff}}_t
=\mathcal{G}(\mathcal{B}_t;\theta_t)$.
Let $\mathcal{U}_{\mathrm{SCC}}$ denote the update induced by Equation~\ref{eq:coupled_opd_objective}, with optimizer state and learning-rate settings left implicit. The update and subsequent realization satisfy:
\begin{equation}
\begin{aligned}
\theta_{t+1}
&=
\mathcal{U}_{\mathrm{SCC}}
\!\left(
\theta_t,\mathcal{B}_t,\mathcal{B}^{\mathrm{eff}}_t;\pi_T
\right),\\
\mathcal{B}^{\mathrm{eff}}_{t+1}
&=
\mathcal{G}
\!\left(\mathcal{B}_{t+1};\theta_{t+1}\right).
\end{aligned}
\label{eq:student_curriculum_coupling}
\end{equation}
The second line applies whenever a subsequent scheduled batch exists. 

Equation~\ref{eq:student_curriculum_coupling} closes the loop between student optimization and supervision realization, while the candidate space and schedule remain fixed.

\vspace{-0.5em}
\section{Experiments}
\vspace{-0.5em}

\begin{table*}[!t]
    \centering
    \small
    \setlength{\tabcolsep}{4.2pt}
    
    \caption{
        \textbf{Evaluation on three TVG benchmarks.}
        Published benchmark results are taken from
        Video-OPD~\citep{li2026video}.
        Bold values indicate the best performance among the non-proprietary
        methods shown in the table.
    }
    \label{tab:main_tvg_results}
    \resizebox{\textwidth}{!}{
    \begin{tabular}{l|ccc|ccc|ccc}
        \toprule
        & \multicolumn{3}{c|}{\textbf{Charades-TimeLens}}
        & \multicolumn{3}{c|}{\textbf{ActivityNet-TimeLens}}
        & \multicolumn{3}{c}{\textbf{QVHighlights-TimeLens}} \\
        \cmidrule(lr){2-4}
        \cmidrule(lr){5-7}
        \cmidrule(lr){8-10}
        \textbf{Models for Evaluation}
        & \textbf{R@0.3} & \textbf{R@0.5} & \textbf{R@0.7}
        & \textbf{R@0.3} & \textbf{R@0.5} & \textbf{R@0.7}
        & \textbf{R@0.3} & \textbf{R@0.5} & \textbf{R@0.7} \\
        \midrule

        \rowcolor{gray!15}
        \multicolumn{10}{l}{\textit{Proprietary Models}} \\

        GPT-4o~\citep{hurst2024gpt4o}
        & 60.6 & 44.5 & 23.5
        & 55.2 & 41.4 & 25.8
        & 69.0 & 54.8 & 38.5 \\

        GPT-5~\citep{singh2025gpt5}
        & 59.3 & 42.0 & 22.0
        & 57.4 & 44.9 & 30.4
        & 72.4 & 60.4 & 46.4 \\

        Gemini-2.0-Flash~\citep{comanici2025gemini}
        & 66.4 & 53.5 & 27.1
        & 62.9 & 54.0 & 37.7
        & 76.2 & 66.4 & 48.3 \\

        Gemini-2.5-Flash~\citep{comanici2025gemini}
        & 68.7 & 56.1 & 30.6
        & 66.8 & 57.5 & 41.3
        & 78.2 & 69.4 & 55.0 \\

        Gemini-2.5-Pro~\citep{comanici2025gemini}
        & 74.1 & 61.1 & 34.0
        & 72.3 & 64.2 & 47.1
        & 84.1 & 75.9 & 61.1 \\

        \midrule
        \rowcolor{gray!15}
        \multicolumn{10}{l}{\textit{Open-Source Models}} \\

        VideoChat-Flash-7B~\citep{li2025videochatflash}
        & 60.2 & 37.9 & 17.8
        & 35.5 & 21.8 & 10.5
        & 45.2 & 30.6 & 16.7 \\

        Qwen2.5-VL-7B~\citep{bai2025qwen25vl}
        & 58.1 & 35.1 & 18.2
        & 47.2 & 32.5 & 20.2
        & 55.0 & 41.7 & 29.3 \\

        VideoChat-R1-7B~\citep{li2025videochatr1}
        & 51.9 & 30.8 & 11.7
        & 35.0 & 23.9 & 11.3
        & 29.3 & 19.1 & 9.4 \\

        Time-R1-7B~\citep{wang2025timer1}
        & 57.9 & 32.0 & 16.9
        & 44.8 & 31.0 & 19.0
        & 65.8 & 51.5 & 36.1 \\

        TVG-R1-7B~\citep{chen2025tvgr1}
        & 44.5 & 23.6 & 12.4
        & 46.7 & 31.0 & 18.6
        & 55.8 & 41.2 & 28.0 \\

        VideoChat-R1.5-7B~\citep{yan2025videochatr15}
        & 46.4 & 24.0 & 10.4
        & 40.6 & 25.3 & 16.4
        & 62.2 & 44.5 & 28.3 \\

        MiMo-VL-7B~\citep{xiaomi2025mimovl}
        & 57.9 & 42.6 & 20.5
        & 49.3 & 38.7 & 22.4
        & 57.1 & 42.6 & 28.4 \\

        Qwen3-VL-8B-Instruct~\citep{bai2025qwen3vl}
        & 61.7 & 41.5 & 23.1
        & 41.2 & 30.7 & 20.0
        & 46.6 & 38.2 & 29.5 \\

        \midrule
        \rowcolor{gray!15}
        \multicolumn{10}{l}{\textit{Post-Training Frameworks}} \\

        OP-RKD [Qwen3-VL-8B]~\citep{li2026video}
        & 67.3 & 47.0 & 27.8
        & 59.1 & 44.9 & 30.6
        & 70.7 & 57.4 & 44.5 \\

        OP-FKD [Qwen3-VL-8B]~\citep{li2026video}
        & 66.9 & 47.3 & 27.6
        & 58.6 & 44.2 & 30.5
        & 71.2 & 58.9 & 46.5 \\

        GRPO [Qwen3-VL-8B]~\citep{li2026video}
        & 72.7 & 44.4 & 27.6
        & 58.6 & 42.7 & 32.1
        & 69.8 & 53.0 & 41.5 \\

        Video-OPD [Qwen3-VL-8B]~\citep{li2026video}
        & 73.1 & 45.8 & 32.4
        & 60.5 & 45.6 & 35.8
        & 73.8 & 60.3 & 50.4 \\

        \textbf{SCC [Qwen3-VL-8B] (ours)}
        & \textbf{74.0} & \textbf{48.1} & \textbf{33.0}
        & \textbf{64.8} & \textbf{49.2} & \textbf{38.2}
        & \textbf{76.8} & \textbf{63.9} & \textbf{54.3} \\

        \bottomrule
    \end{tabular}
    }
    \vspace{-0.8em}
\end{table*}

Section~\ref{sec:experimental_setup} describes the training setup, AF curriculum construction, and evaluation protocol. Section~\ref{sec:main_results} reports the main accuracy and efficiency results, while Section~\ref{sec:ablations} decomposes SCC by separately evaluating candidate-space composition and student-dependent realization. Additional results and implementation details are provided in Appendix~\ref{app:implementation_experiments}.

\vspace{-0.5em}
\subsection{Experimental Setup}
\label{sec:experimental_setup}

\paragraph{Training Details.}

We use Qwen3-VL-8B-Instruct~\citep{bai2025qwen3vl} as the student. For direct comparison with Video-OPD~\citep{li2026video}, we use its released Qwen3-VL-32B teacher, trained with group relative policy optimization (GRPO). Following the same configuration, we set the maximum input video token length to 8,192, sample videos at 2 FPS, cap the number of frames at 768, and set the maximum video-frame token length to 768. We use a learning rate of $1 \times 10^{-6}$ and a 79-step training schedule with variable scheduled global batch sizes capped at 32 (Appendix~\ref{app:curriculum_details}). Each scheduled example uses a single on-policy rollout, and we follow Video-OPD for rollout sampling, video preprocessing, and the remaining optimization settings. We set $\tau_S=0.7$ for student-dependent realization throughout the main experiments; sensitivity to this criterion is examined in Appendix~\ref{app:student_criterion}.

\vspace{-0.5em}
\paragraph{AF curriculum construction.} Following the AF design in Section~\ref{sec:af_candidate_space}, we construct the candidate supervision space from TimeLens-100K~\citep{zhang2025timelens}, using public data from HiREST~\citep{zala2023hirest}, QuerYD~\citep{oncescu2021queryd}, CosMo-Cap~\mbox{\citep{wang2024cosmo}}, InternVid-VTime~\citep{wang2024internvid,huang2024vtimellm}, and DiDeMo~\citep{hendricks2017didemo}. We first retain teacher-qualified examples with $q_T(x)\geq0.7$, and then select 100 Anchor examples satisfying $q_{S,0}(x)\geq0.7$ and 900 Frontier examples satisfying $q_{S,0}(x)<0.3$. The resulting 1,000-example curriculum is assigned to a 79-step training schedule. We exclude duplicate annotations and videos overlapping with the TVG evaluation sets. Source composition and training-schedule details are provided in Appendix~\ref{app:curriculum_details}, with sensitivity to the Anchor--Frontier ratio examined in Appendix~\ref{app:af_ratio}.


\vspace{-0.5em}
\paragraph{Evaluation Benchmarks and Metrics.}
Our evaluation focuses on TVG performance and training efficiency. We evaluate on Charades-STA~\citep{gao2017tall}, ActivityNet~\citep{caba2015activitynet}, and QVHighlights~\citep{lei2021qvhighlights}, using the corrected temporal annotations from TimeLens~\citep{zhang2025timelens}. Following prior work~\citep{li2025imove,wang2025grounded,wang2024hawkeye}, we report mean IoU (mIoU) and Recall at IoU thresholds of $0.3$, $0.5$, and $0.7$. We assess training efficiency by curriculum size, the number of teacher-scored OPD routes, and training time.

\vspace{-0.4em}
\subsection{Main Results}
\label{sec:main_results}


\paragraph{Temporal Video Grounding.} Table~\ref{tab:main_tvg_results} compares SCC with representative proprietary models, open-source VLMs, and post-training frameworks across three TVG benchmarks. Our method achieves the best performance among the non-proprietary methods in Table~\ref{tab:main_tvg_results} across all reported recall metrics and consistently outperforms Video-OPD. Comparison with TimeLens-8B reveals benchmark-dependent trade-offs under different training regimes (Appendix~\ref{app:timelens_comparison}).

\vspace{-0.5em}

\begin{figure}[ht]
    \centering
    \setlength{\abovecaptionskip}{4pt}
    \includegraphics[width=\linewidth]{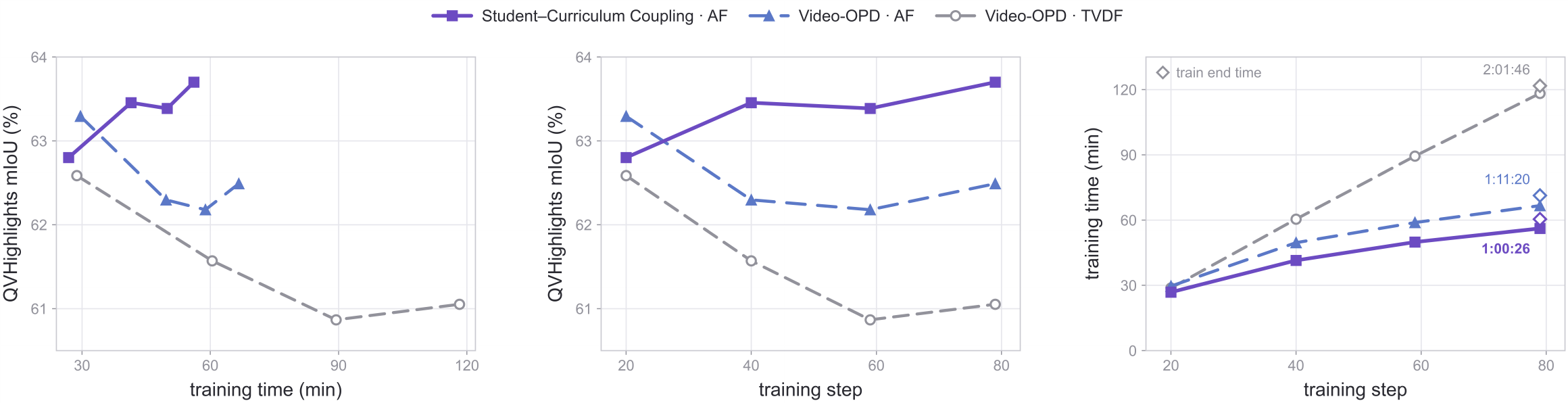}
    \vspace{-1em}
    \caption{
    \textbf{Accuracy--efficiency comparison.}
    QVHighlights mIoU versus training time (left), QVHighlights mIoU across training steps (middle), and cumulative training time (right).
    Diamonds mark training completion.
    All times are measured on eight NVIDIA A100 GPUs.
    }
    \label{fig:training_efficiency}
    \vspace{-0.8em}
\end{figure}

\paragraph{Accuracy and Training Efficiency.} Table~\ref{tab:accuracy_efficiency} reports curriculum scale and supervision routing, while Figure~\ref{fig:training_efficiency} summarizes accuracy and training time. For comparison, we additionally apply Video-OPD to the same AF curriculum used by SCC. With the same 1,000 examples and 1,004 scheduled rollouts, SCC requires only 492 teacher-scored OPD routes, a $51.0\%$ reduction from Video-OPD. With AF held fixed, replacing uniform OPD with student-dependent routing yields an approximately 2.1\% relative improvement in mean recall and reduces training time by 15.3\%. Compared with Video-OPD on TVDF, SCC uses 60.0\% fewer curriculum examples and achieves higher endpoint mIoU in approximately half the training time.

\begin{table*}[ht]
\centering
\scriptsize
\setlength{\tabcolsep}{2.7pt}
\renewcommand{\arraystretch}{1.08}

\caption{
\textbf{Training data and supervision routing.}
All runs use 79 scheduled training steps. Presentations / Rollouts counts student rollouts; Teacher / OPD Routes counts rollouts receiving teacher scoring and OPD optimization; Suspended Routes counts rollouts not routed to teacher scoring or OPD after current-student assessment.
}
\label{tab:accuracy_efficiency}

\begin{tabular*}{\textwidth}{
    @{\extracolsep{\fill}}
    llrrrrr
    @{}
}
\toprule

\multirow[c]{2}{*}{
    \raisebox{-0.3ex}{\textbf{Method}}
}
&
\multirow[c]{2}{*}{
    \raisebox{-0.3ex}{\textbf{Curriculum}}
}
&
\multicolumn{2}{c}{\textbf{Data and Schedule}}
&
\multicolumn{3}{c}{\textbf{Supervision Routing}}
\\

\cmidrule(lr){3-4}
\cmidrule(lr){5-7}

&
&
\textbf{\# Examples}
&
\shortstack{
    \textbf{Presentations}\\[-0.2ex]
    \textbf{/ Rollouts}
}
&
\shortstack{
    \textbf{Teacher / OPD}\\[-0.2ex]
    \textbf{Routes}
}
&
\shortstack{
    \textbf{Suspended}\\[-0.2ex]
    \textbf{Routes}
}
&
\textbf{Route Rate}
\\

\midrule

Video-OPD
& TVDF
& 2,500
& 2,504
& 2,504
& 0
& 100.0\%
\\

Video-OPD
& AF
& 1,000
& 1,004
& 1,004
& 0
& 100.0\%
\\

\textbf{SCC}
& \textbf{AF}
& \textbf{1,000}
& \textbf{1,004}
& \textbf{492}
& \textbf{512}
& \textbf{49.0\%}
\\

\bottomrule
\end{tabular*}
\vspace{-0.8em}
\end{table*}

\vspace{-0.4em}
\subsection{Decomposing Student--Curriculum Coupling}
\label{sec:ablations}

SCC combines a capability-structured candidate space with student-dependent realization of its supervision. Table~\ref{tab:scc_ablations} decomposes their contributions by varying one component while holding the other fixed. All trained variants share the same student initialization and frozen teacher and are trained for 79 scheduled steps.

\begin{table*}[t]
    \centering
    \scriptsize
    \setlength{\tabcolsep}{2.7pt}
    \caption{
    \textbf{Decomposition of Student--Curriculum Coupling.} Base Model denotes Qwen3-VL-8B-Instruct before post-training, with scores taken from Video-OPD~\citep{li2026video}. All other results are obtained using the same local evaluation pipeline. The upper block varies the candidate space with student-dependent realization fixed; the lower block fixes AF and compares student-dependent realization with uniform OPD supervision. TVDF uses the original 2,500-example curriculum from Video-OPD~\citep{li2026video}, while Random, Frontier-only, and AF each use 1,000 examples. $\checkmark$ denotes student-dependent realization, and $\times$ denotes uniform OPD supervision over all scheduled examples. \textsuperscript{$\star$} marks benchmarks re-annotated with TimeLens. Values are percentages rounded to one decimal place; bold marks the best result for each metric within each block.
}
    \label{tab:scc_ablations}
    \resizebox{\textwidth}{!}{
    \begin{tabular}{lc|rrrr|rrrr|rrrr}
        \toprule
        \textbf{Candidate}
        & \textbf{Student-Dep.}
        & \multicolumn{4}{c|}{
            \textbf{Charades\textsuperscript{$\star$}}
        }
        & \multicolumn{4}{c|}{
            \textbf{ActivityNet\textsuperscript{$\star$}}
        }
        & \multicolumn{4}{c}{
            \textbf{QVHighlights\textsuperscript{$\star$}}
        } \\
        \textbf{   Space}
        & \textbf{Realization}
        & \textbf{mIoU}
        & \textbf{R@0.3}
        & \textbf{R@0.5}
        & \textbf{R@0.7}
        & \textbf{mIoU}
        & \textbf{R@0.3}
        & \textbf{R@0.5}
        & \textbf{R@0.7}
        & \textbf{mIoU}
        & \textbf{R@0.3}
        & \textbf{R@0.5}
        & \textbf{R@0.7} \\
        \midrule

        \rowcolor{gray!8}
        Base Model
        & --
        & 42.9 & 61.7 & 41.5 & 23.1
        & 30.4 & 41.2 & 30.7 & 20.0
        & 36.9 & 46.6 & 38.2 & 29.5 \\

        \midrule
        \rowcolor{gray!15}
        \multicolumn{14}{l}{
            \textit{(a) Candidate-Space Composition}
        } \\

        TVDF
        & $\checkmark$
        & 52.1 & 72.7 & 46.2 & 32.9
        & 49.3 & 62.1 & 47.4 & 37.4
        & 61.2 & 73.6 & 60.2 & 50.9 \\

        Random
        & $\checkmark$
        & 51.6 & 73.0 & 45.9 & 31.7
        & 48.8 & 62.3 & 46.4 & 36.4
        & 60.9 & 73.8 & 59.8 & 50.0 \\

        Frontier-only
        & $\checkmark$
        & 52.0 & 73.1 & 46.3 & 32.5
        & 49.5 & 63.1 & 47.4 & 37.3
        & 62.3 & 75.3 & 62.0 & 51.9 \\

        \textbf{AF}
        & $\checkmark$
        & \textbf{52.8}
        & \textbf{74.0}
        & \textbf{48.1}
        & \textbf{33.0}
        & \textbf{50.6}
        & \textbf{64.8}
        & \textbf{49.2}
        & \textbf{38.2}
        & \textbf{63.7}
        & \textbf{76.8}
        & \textbf{63.9}
        & \textbf{54.3} \\

        \midrule
        \rowcolor{gray!15}
        \multicolumn{14}{l}{
            \textit{(b) Student-Dependent Realization}
        } \\

        AF
        & $\times$
        & 52.3 & 73.3 & 47.1 & 32.8
        & 49.6 & 63.4 & 47.8 & 37.2
        & 62.5 & 75.3 & 62.6 & 52.1 \\

        \textbf{AF}
        & $\checkmark$
        & \textbf{52.8}
        & \textbf{74.0}
        & \textbf{48.1}
        & \textbf{33.0}
        & \textbf{50.6}
        & \textbf{64.8}
        & \textbf{49.2}
        & \textbf{38.2}
        & \textbf{63.7}
        & \textbf{76.8}
        & \textbf{63.9}
        & \textbf{54.3} \\

        \bottomrule
    \end{tabular}
    }
    \vspace{-0.8em}
\end{table*}

\vspace{-0.5em}
\paragraph{Candidate-Space Composition.} Table~\ref{tab:scc_ablations} (a) compares candidate curricula under the same student-dependent realization. AF performs best across all metrics and benchmarks. Its advantage over the matched-size Random curriculum supports capability-structured selection, while the comparison with Frontier-only highlights the complementary role of Anchors. AF also outperforms the larger TVDF curriculum, indicating that student-dependent realization benefits from a well-structured candidate space. Sensitivity to the Anchor--Frontier ratio and candidate-space scale is examined in Appendices~\ref{app:af_ratio} and~\ref{app:af_scale}, respectively.

\vspace{-0.5em}
\paragraph{Student-Dependent Realization.}
Table~\ref{tab:scc_ablations} (b) fixes the AF candidate space and removes student dependence from curriculum realization, applying OPD supervision uniformly across scheduled examples. This change reduces every reported metric across all three benchmarks. The consistent decline shows that candidate-space composition alone is insufficient and that effective coupling requires supervision to respond to the student's current competence. Further analyses on TVDF and static routing are provided in Appendices~\ref{app:tvdf_student_realization} and~\ref{app:realization_controls}, respectively.

\vspace{-0.5em}

\section{Training Dynamics of Student--Curriculum Coupling}
\label{sec:training_dynamics}
\vspace{-0.5em}

\begin{figure}[ht]
    \centering
    \setlength{\abovecaptionskip}{4pt}
    \includegraphics[width=\linewidth]{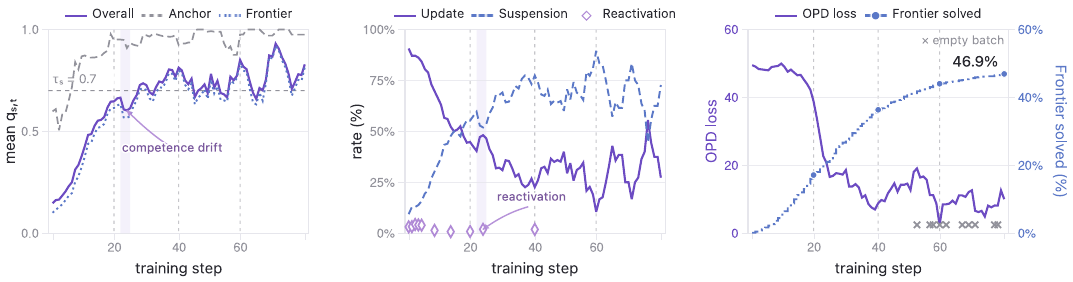}
    \vspace{-1em}
    \caption{
    \textbf{Training dynamics of Student--Curriculum Coupling.}
    Current student competence for Anchors, Frontiers, and the full AF
    curriculum (left); realized OPD updates, suspensions, and Anchor
    reactivations (middle); and effective-batch OPD loss with cumulative
    Frontier acquisition (right).
    The shaded interval marks a representative drop in Anchor competence and the
    resulting reactivation of supervision.
    }
    \label{fig:training_dynamics}
    \vspace{-0.8em}
\end{figure}

Figure~\ref{fig:training_dynamics} traces the coupled evolution of student competence and the effective curriculum, highlighting the complementary roles of Anchors and Frontiers during training. Anchor competence remains high in the early stage, and most Anchor presentations are suspended, limiting additional optimization in already-supported capability regions. When an Anchor falls below the task criterion, its supervision is reactivated, allowing Anchors to support capability stabilization. Frontiers, meanwhile, account for most active OPD routes, and their competence rises with overall student progress. The AF curriculum therefore combines stabilization of existing capabilities with acquisition along the learning frontier.


As more Frontier examples reach the criterion, cumulative acquisition rises to 46.9\% of the 900 unique Frontier examples. Over the same period, the update rate declines, suspension becomes more prevalent, and effective-batch OPD loss decreases through the main acquisition phase. Later in training, OPD supervision becomes increasingly sparse, with occasional empty effective batches and updates concentrated on the remaining deficits. This progression reflects the closed loop underlying SCC, in which the current student shapes the effective curriculum, the resulting OPD updates alter the student, and the updated state governs supervision in subsequent batches.

\vspace{-0.5em}
\section{Related Work}
\label{sec:related_work}
\vspace{-0.5em}

\paragraph{VLMs for Temporal Video Grounding.}
VLM-based TVG research has focused on both training strategies and temporal modeling. Training approaches range from boundary-aware instruction tuning and time-aware objectives~\citep{huang2024vtimellm,wang2024hawkeye} to reinforcement-based post-training for temporal localization and spatiotemporal perception~\citep{wang2025timer1,li2025videochatr1,chen2025tvgr1}. Complementary work makes temporal structure explicit through timestamp-aware visual encoding~\citep{ren2024timechat} and dedicated temporal representations~\citep{huang2024lita,wang2025grounded}, with TRACE modeling events as sequences of timestamps, saliency scores, and captions~\citep{guo2025trace}. To handle longer videos, TimeSuite combines efficient token processing with grounded tuning~\citep{zeng2025timesuite}, while ReVisionLLM adopts recursive coarse-to-fine localization~\citep{hannan2025revisionllm}. Alongside these modeling advances, TimeLens examines annotation quality and training strategies, providing curated training data and re-annotated benchmarks for more reliable evaluation~\citep{zhang2025timelens}.

\vspace{-0.5em}
\paragraph{On-Policy Distillation.}

OPD trains on student-generated sequences with teacher feedback at the same visited prefixes, reducing distribution mismatch between training and inference~\citep{agarwal2024onpolicy,lu2025onpolicydistillation}. MiniLLM adopts reverse KL to avoid overestimating low-probability teacher outputs~\citep{gu2024minillm}, while entropy-aware OPD supplements this objective with forward KL when teacher entropy is high to preserve generation diversity~\citep{jin2026eopd}. Alongside objective design, OPD analyses examine how compatible reasoning patterns and teacher capabilities novel to the student affect knowledge transfer, linking successful distillation to progressive alignment on high-probability tokens~\citep{li2026rethinking}. Video-OPD extends this paradigm to TVG, providing dense token-level supervision by scoring student-generated trajectories without separate teacher rollouts~\citep{li2026video}. Building on this formulation, our work identifies the implicit persistent-value assumption and develops a coupled view of OPD.

\vspace{-0.5em}
\paragraph{Curriculum Learning and Data Selection.}

Curriculum learning organizes training examples by difficulty or learning structure~\citep{bengio2009curriculum}, while classical self-paced learning adapts an easy-to-hard curriculum alongside model updates~\citep{kumar2010selfpaced}. Online hard example mining and Selective-Backprop instead prioritize high-loss examples~\citep{shrivastava2016training,jiang2019accelerating}. For language models, DoReMi learns domain-mixture weights through proxy-model optimization~\citep{xie2023doremi}, while RegMix predicts effective mixtures from smaller-scale experiments~\citep{liu2025regmix}. Skill-It adapts sampling to skill dependencies and current learning progress~\citep{chen2023skillit}. In TVG, curriculum design encompasses temporal-task complexity and difficulty-aware data construction for reinforcement post-training~\citep{wang2025grounded,chen2025tvgr1,wang2025timer1}. Video-OPD combines teacher validation with teacher--student disagreement for candidate selection~\citep{li2026video}. Compared with prior work on sample prioritization, SCC couples supervision trustworthiness with evolving student competence, jointly designing curriculum composition and supervision realization in OPD for VLM-based TVG.

\vspace{-0.5em}
\section{Conclusion}
\label{sec:conclusion}
\vspace{-0.5em}

This work shows how supervision aligned with evolving task competence can improve both accuracy and efficiency in VLM post-training. Distinguishing supervision trustworthiness from necessity leads to a coupled view of OPD, linking curriculum composition to the student's changing learning needs. SCC realizes this view through a compact AF-defined space of trustworthy supervision opportunities, with current student competence guiding their selective activation. Across three TVG benchmarks, it achieves higher grounding accuracy than Video-OPD on TVDF with fewer training examples, fewer teacher-scored trajectories, and shorter training time. Together with ablations and training-dynamics analyses, these results highlight the value of jointly designing curriculum composition and supervision timing.

Despite these gains, SCC assesses supervision necessity using tIoU between a single sampled prediction and the ground-truth interval. This task-level proxy does not directly estimate the marginal effect of supervising an example on generalization, and routing decisions may be sensitive to sampling variability. Extending this annotation-dependent assessment to tasks without comparable ground truth also requires alternative signals. Beyond these assessment limitations, our evaluation focuses primarily on TVG post-training within one VLM family. Future work should develop more direct, less annotation-dependent estimates of supervision value and evaluate this coupled design across additional model families and tasks.

\vspace{-0.5em}
\clearpage
\subsection*{AI use statement}
\vspace{-0.5em}

Generative AI tools assisted with manuscript refinement, discussions of derivations and proof arguments, feedback on methodology and experimental design, and interpretation of results. The authors remain responsible for all scientific claims, references, experimental results, and other content of the manuscript.
\vspace{-0.5em}

\subsection*{Ethics statement}
\vspace{-0.5em}

This work studies temporal video grounding using existing video datasets and pretrained models described in Section~\ref{sec:experimental_setup}. The method may inherit biases and limitations from these resources. Improved video localization can support retrieval and accessibility but may also facilitate privacy-invasive video analysis. Deployment should account for consent, privacy, dataset bias, and localization errors, particularly in applications involving people. Use or redistribution of the data and models remains subject to their original licenses and access conditions. The reported benchmark results do not establish suitability for safety-critical deployment.
\vspace{-0.5em}

\subsection*{Reproducibility statement}
\vspace{-0.5em}

Sections~\ref{sec:method} and~\ref{sec:experimental_setup} describe the method, model configurations, video preprocessing, training settings, and evaluation protocol. Appendix~\ref{app:theory} provides supporting derivations, and Appendix~\ref{app:implementation_experiments} details the algorithm, curriculum construction, training schedule, and supplementary experiments.

\vspace{-0.5em}

\subsection*{Acknowledgments}
\vspace{-0.5em}

We acknowledge the creators of TimeLens-100K and TimeLens-Bench~\citep{zhang2025timelens}, which are released under the License Term of TimeLens. These resources are used solely for academic research, with no commercial or production use. We also thank the contributors to the underlying video datasets for making these resources available.

\bibliography{iclr2027_conference}
\bibliographystyle{iclr2027_conference}

\clearpage
\appendix
\section*{Appendix}
\vspace{-0.5em}
\section{Theoretical Analysis}
\label{app:all}
\label{app:theory}
\vspace{-0.5em}
This appendix develops the theoretical and optimization analysis
underlying SCC.
Section~\ref{app:necessity_value} analyzes the state dependence of
marginal supervision value and distinguishes task-criterion satisfaction
from distributional alignment.
Section~\ref{app:opd_estimator} derives the sampled-token OPD estimator,
and Section~\ref{app:normalization_execution} formalizes
student-dependent realization, objective normalization, and selective
execution.

\vspace{-0.5em}
\subsection{Supervision Necessity and Marginal Value}
\label{app:necessity_value}

\vspace{-0.4em}
\paragraph{Local marginal effects.}
Using the paired updates defining $\Delta_t(x)$, let:
\begin{equation}
d_t(x)
=
\theta_{t+1}^{+x}-\theta_{t+1}^{-x}.
\label{eq:supervision_update_displacement}
\end{equation}
The two updates share all other batch contributions, normalization,
initial optimizer state, and training randomness.
Thus, $d_t(x)$ captures the optimizer's response to including the
on-policy distillation (OPD) contribution of $x$.
To expose the state dependence of supervision value, we relate this
displacement to the local change in population temporal video grounding
(TVG) performance.

Suppose $J_{\mathrm{TVG}}$ has an $L_J$-Lipschitz gradient on a common
convex neighborhood containing all paired updates, and the expectations
below are finite.
The following expansion is conditional on this regularity assumption,
which need not hold for metrics based on discrete decoding.
A first-order expansion gives:
\begin{equation}
\begin{aligned}
\Delta_t(x)
&=
\mathbb{E}\!\left[
\left\langle
\nabla J_{\mathrm{TVG}}(\theta_{t+1}^{-x}),
d_t(x)
\right\rangle
\,\middle|\,\theta_t,x
\right]
+R_t(x),\\
|R_t(x)|
&\leq
\frac{L_J}{2}
\mathbb{E}\!\left[
\|d_t(x)\|_2^2
\,\middle|\,\theta_t,x
\right].
\end{aligned}
\label{eq:local_marginal_supervision_value}
\end{equation}
Here, $R_t(x)$ is the conditional expected first-order remainder.
For each paired update, integrating the gradient along
$\theta_{t+1}^{-x}+s\,d_t(x)$ for $s\in[0,1]$ and applying Lipschitz
continuity bounds the remainder by $L_J\|d_t(x)\|_2^2/2$ in absolute
value.
Taking conditional expectations yields the stated bound.

The leading term measures the alignment of the supervision-induced
displacement with the local direction of increasing population
performance.
In OPD, this displacement reflects the student's sampled prefixes,
the corresponding distillation signals, and the optimizer state.
Both the displacement and the performance gradient can vary as the
student evolves, allowing supervision from a fixed teacher on a fixed
example to have changing marginal effects.

\vspace{-0.5em}
\paragraph{Task-criterion satisfaction and distributional alignment.}
A realized prediction can satisfy the task-level criterion while
distributional alignment remains incomplete.
Consider a one-step model with two output symbols encoding a correct
interval $y_c$ and an incorrect interval $y_w$.
Let the student and teacher distributions be
$p_S=(\alpha,1-\alpha)$ and $p_T=(\beta,1-\beta)$, respectively, with
$\alpha,\beta\in(1/2,1)$ and $\alpha\neq\beta$.
For $\tau_S\leq1$, a student rollout producing $y_c$ has temporal
intersection over union (tIoU) equal to one and satisfies the task-level
competence criterion used for routing.
Nevertheless,
\begin{equation}
\begin{aligned}
D_{\mathrm{KL}}(p_S\|p_T)
&=
\alpha\log\frac{\alpha}{\beta}
+(1-\alpha)\log\frac{1-\alpha}{1-\beta}
>0,\\
\frac{\partial}{\partial\alpha}
D_{\mathrm{KL}}(p_S\|p_T)
&=
\log\frac{\alpha(1-\beta)}{\beta(1-\alpha)}
\neq0.
\end{aligned}
\label{eq:task_alignment_separation}
\end{equation}
Thus, satisfying the task criterion need not eliminate the
distributional alignment signal.
The criterion evaluates the realized task outcome, while the
Kullback--Leibler (KL) divergence compares predictive distributions.

\vspace{-0.5em}
\paragraph{Implications for supervision allocation.}
These observations motivate separating the availability of trustworthy
supervision from its realization during training.
Initial suitability alone does not guarantee persistent marginal value,
and a residual distillation discrepancy does not establish a deficit
in the realized task outcome.
In Student--Curriculum Coupling (SCC), the necessity indicator uses
current task performance to guide supervision activation without
directly estimating $\Delta_t(x)$.
Within this allocation principle, the Anchor--Frontier (AF) space
retains trustworthy opportunities for capability acquisition and
stabilization.

\vspace{-0.4em}
\subsection{Sampled-Token OPD Estimator}
\label{app:opd_estimator}

Section~\ref{sec:opd} defines token-level reverse-KL alignment under a
fixed rollout distribution.
Here, we present the sampled-token surrogate used for
optimization~\citep{li2026video} and establish its local gradient
identity with Equation~\ref{eq:opd_objective}.
The derivation assumes bounded rollout lengths and smooth, positive
token probabilities over a finite vocabulary.

\vspace{-0.5em}
\paragraph{Sampled-token supervision.}
At training step $t$, the rollout policy $\pi_{\theta_t}$ generates:
\[
y_t=(y_{t,1},\ldots,y_{t,L_t})
\sim\pi_{\theta_t}(\cdot\mid x),
\]
with prefix $h_{t,k}=(x,y_{t,<k})$ at position $k$.
For a vocabulary token $z$, define:
\[
p_{t,k}(z) = \pi_{\theta_t}(z \mid h_{t,k}),
\qquad
q_{t,k}(z) = p_{T,k}(z) = \pi_T(z \mid h_{t,k}),
\]
the rollout-policy and frozen-teacher probabilities at the same prefix.
The sampled-token supervision signal is:
\begin{equation}
r_{t,k}
=
\log q_{t,k}(y_{t,k})
-
\log p_{t,k}(y_{t,k}).
\label{eq:sampled_opd_signal}
\end{equation}
Since $y_{t,k}$ is sampled from $p_{t,k}$ conditional on $h_{t,k}$,
\begin{equation}
\mathbb{E}_{y_{t,k}\sim p_{t,k}}
\left[-r_{t,k}\mid h_{t,k}\right]
=
D_{\mathrm{KL}}
\!\left(p_{t,k}\,\middle\|\,q_{t,k}\right).
\label{eq:sampled_opd_unbiased}
\end{equation}
Thus, $-r_{t,k}$ is an unbiased single-token estimate of the reverse-KL
value at that prefix~\citep{li2026rethinking}.

\vspace{-0.5em}
\paragraph{Importance-weighted surrogate.}
During the update, $\theta_t$ remains fixed, while $\theta$ denotes the
optimization variable initialized from $\theta_t$.
The token-level likelihood ratio is:
\begin{equation}
\rho_{t,k}(\theta)
=
\frac{\pi_\theta(y_{t,k}\mid h_{t,k})}
{\pi_{\theta_t}(y_{t,k}\mid h_{t,k})}.
\label{eq:opd_likelihood_ratio}
\end{equation}
Following Video-OPD~\citep{li2026video}, the per-example surrogate is:
\begin{equation}
\ell_{\mathrm{OPD}}
\!\left(x,y_t;\theta,\theta_t,\pi_T\right)
=
-\sum_{k=1}^{L_t}
\operatorname{sg}[r_{t,k}]\,
\rho_{t,k}(\theta),
\label{eq:sampled_opd_surrogate}
\end{equation}
where $\operatorname{sg}[\cdot]$ denotes stop-gradient.
Holding the sampled trajectory, rollout probabilities, and supervision
signals fixed gives:
\begin{equation}
\nabla_\theta\ell_{\mathrm{OPD}}
=
-\sum_{k=1}^{L_t}
\operatorname{sg}[r_{t,k}]\,
\rho_{t,k}(\theta)\,
\nabla_\theta
\log\pi_\theta(y_{t,k}\mid h_{t,k}).
\label{eq:sampled_opd_gradient}
\end{equation}
At the rollout point $\theta = \theta_t$, the likelihood ratio equals one, and its gradient satisfies:
\[
\left.
\nabla_\theta\rho_{t,k}(\theta)
\right|_{\theta=\theta_t}
=
\left.
\nabla_\theta\log\pi_\theta(y_{t,k}\mid h_{t,k})
\right|_{\theta=\theta_t}.
\]

\vspace{-0.5em}
\paragraph{Relation to the fixed-rollout objective.}
For a fixed prefix $h_{t,k}$, write
$p_{\theta,k}(z)=\pi_\theta(z\mid h_{t,k})$, so that
$p_{\theta_t,k}=p_{t,k}$.
Differentiating the token-level reverse KL yields:
\begin{equation}
\begin{aligned}
\nabla_\theta
D_{\mathrm{KL}}
\!\left(p_{\theta,k}\,\middle\|\,q_{t,k}\right)
&=
\mathbb{E}_{z\sim p_{\theta,k}}
\left[
\left(
\log\frac{p_{\theta,k}(z)}{q_{t,k}(z)}+1
\right)
\nabla_\theta\log p_{\theta,k}(z)
\right]\\
&=
\mathbb{E}_{z\sim p_{\theta,k}}
\left[
\log\frac{p_{\theta,k}(z)}{q_{t,k}(z)}
\nabla_\theta\log p_{\theta,k}(z)
\right].
\end{aligned}
\label{eq:reverse_kl_gradient}
\end{equation}
The second equality follows from the score-function identity
$\mathbb{E}_{z\sim p_{\theta,k}}
[\nabla_\theta\log p_{\theta,k}(z)]=0$.
Combining Equations~\ref{eq:sampled_opd_signal},
\ref{eq:sampled_opd_gradient}, and~\ref{eq:reverse_kl_gradient}
at $\theta=\theta_t$ gives:
\begin{equation}
\begin{aligned}
&\left.
\mathbb{E}_{y_{t,k}\sim p_{t,k}}
\left[
\nabla_\theta
\left(-\operatorname{sg}[r_{t,k}]\rho_{t,k}(\theta)\right)
\,\middle|\,h_{t,k}
\right]
\right|_{\theta=\theta_t}\\
&\qquad =
\left.
\nabla_\theta
D_{\mathrm{KL}}
\!\left(p_{\theta,k}\,\middle\|\,q_{t,k}\right)
\right|_{\theta=\theta_t}.
\end{aligned}
\label{eq:opd_gradient_equivalence}
\end{equation}
Taking expectation over the fixed rollout policy and summing over
generated positions establishes the connection to the objective in
Section~\ref{sec:opd}:
\begin{equation}
\begin{aligned}
&\mathbb{E}_{y_t\sim\pi_{\theta_t}(\cdot\mid x)}
\left[
\left.
\nabla_\theta
\ell_{\mathrm{OPD}}
\!\left(x,y_t;\theta,\theta_t,\pi_T\right)
\right|_{\theta=\theta_t}
\right]\\
&\qquad =
\left.
\nabla_\theta
\mathcal{L}_{\mathrm{OPD},t}(\theta;x)
\right|_{\theta=\theta_t}.
\end{aligned}
\label{eq:opd_fixed_rollout_gradient}
\end{equation}
This identity establishes gradient matching at $\theta=\theta_t$ with
the rollout distribution fixed during differentiation.
The full sequence-level reverse KL additionally differentiates prefix
visitation probabilities, yielding long-horizon score-function
terms~\citep{gu2024minillm}.
Section~\ref{sec:closed_loop_optimization} aggregates the per-example
surrogate above over the effective batch with fixed routing decisions.

\vspace{-0.4em}
\subsection{Objective Normalization and Efficient Execution}
\label{app:normalization_execution}
\label{app:realization_details}

The gradient identity in Appendix~\ref{app:opd_estimator} applies to
the base OPD surrogate before task-dependent routing.
SCC optimizes the masked surrogate in Equation~\ref{eq:coupled_opd_objective}, with routing decisions held
fixed during each update.
Because retention depends on the sampled trajectory, it changes both
the frequency of supervision and the trajectory gradients contributing
to the expected update.

Anchor and Frontier membership is fixed by the initial student assessment.
For presentation $i$ at step $t$, the current trajectory $y_{t,i}$
determines the supervision mask:
\begin{equation}
m_{t,i}
=
\mathbf{1}\!\left\{
q_{S,t}(x_i;y_{t,i})<\tau_S
\right\}.
\label{eq:realized_supervision_mask}
\end{equation}
The task score uses the ground-truth interval, and token-level signals
are teacher-derived.
Scores at or above $\tau_S$ suspend supervision.
Retained trajectories are reused for teacher scoring and OPD without
additional sampling.

The same rule applies to both curriculum regions.
Retained Frontiers receive acquisition-oriented supervision.
An Anchor $x_i\in\mathcal{A}$ is reactivated when
$q_{S,0}(x_i)\geq\tau_S$ and $m_{t,i}=1$.
Anchor selection uses $\tau_A$; reactivation compares initial and
current performance against $\tau_S$ and can occur on the first
scheduled presentation.
Acquisition and reactivation both enter the effective batch and the
supervised count.
Routing statistics count supervised and suspended presentations,
including repeated occurrences; curriculum size counts unique examples.


For a nonempty scheduled batch, let $n_t$ denote the preset normalization denominator and $M_t=\sum_{i=1}^{|\mathcal{B}_t|}m_{t,i}$ the number of supervised presentations. Let $\ell_{t,i}(\theta)$ abbreviate the per-example surrogate $\ell_{\mathrm{OPD}}(x_i,y_{t,i};\theta,\theta_t,\pi_T)$ defined in
Equation~\ref{eq:sampled_opd_surrogate}.
The coupled objective in Equation~\ref{eq:coupled_opd_objective} has the
equivalent masked form:
\begin{equation}
\mathcal{L}^{\mathrm{SCC}}_t(\theta)
=
\frac{1}{n_t}
\sum_{i=1}^{|\mathcal{B}_t|}m_{t,i}\ell_{t,i}(\theta)
=
\frac{1}{n_t}
\sum_{i:m_{t,i}=1}\ell_{t,i}(\theta).
\label{eq:masked_scc_objective}
\end{equation}
With the sampled trajectories, teacher signals, and masks held fixed
during the update, its gradient is:
\begin{equation}
\nabla_\theta\mathcal{L}^{\mathrm{SCC}}_t(\theta)
=
\frac{1}{n_t}
\sum_{i:m_{t,i}=1}\nabla_\theta\ell_{t,i}(\theta).
\label{eq:masked_scc_gradient}
\end{equation}
Thus, evaluating only the retained terms is algebraically equivalent to
masking the full scheduled batch, provided that the denominator remains
$n_t$.
Teacher scores and OPD losses for suspended trajectories need not be
computed.

For $M_t>0$, define the active-only mean as
$\mathcal{L}^{\mathrm{active}}_t(\theta)
=M_t^{-1}\sum_{i:m_{t,i}=1}\ell_{t,i}(\theta)$.
Then:
\begin{equation}
\mathcal{L}^{\mathrm{SCC}}_t(\theta)
=
\frac{M_t}{n_t}\,
\mathcal{L}^{\mathrm{active}}_t(\theta).
\label{eq:scc_normalization_relation}
\end{equation}
Active-only normalization would increase each retained term's weight by $n_t/M_t$ relative to the coupled objective. Normalization by $n_t$ preserves the weight $1/n_t$, so suspension removes supervision contributions without amplifying the remaining terms. The total retained weight is $M_t/n_t$. This scaling applies to the objective and its gradient, but does not imply proportional scaling of parameter updates under adaptive optimization.

When $M_t=0$, the masked objective and its gradient are zero.
The optimizer and weight-decay updates are explicitly skipped, preserving
the student parameters and optimizer state while the training schedule
advances.
All scheduled presentations still require student generation and
assessment; selective execution removes only the subsequent teacher
scoring and OPD computation for suspended trajectories.


\section{Implementation Details and Additional Experiments}
\label{app:implementation_experiments}
\vspace{-0.5em}

Sections~\ref{app:algorithm} and~\ref{app:curriculum_details} describe
the SCC algorithm, AF construction, and training schedule.
The remaining subsections report supplementary accuracy--efficiency
results and analyses of alternative objectives, student-dependent
realization beyond AF, competence-criterion sensitivity, teacher choice,
checkpoint-wise performance, the Anchor--Frontier ratio, and broader
video understanding.

\vspace{-0.5em}
\subsection{Algorithm}
\label{app:algorithm}
\vspace{-0.5em}
Algorithm~\ref{alg:student_curriculum_coupling} summarizes SCC, from AF construction to
student-dependent routing and selective OPD updates.

\vspace{-0.5em}
\begin{algorithm}[ht]
\caption{Student--Curriculum Coupling (SCC)}
\label{alg:student_curriculum_coupling}
\begin{algorithmic}[1]

\Require Candidate pool $\mathcal{P}$, initial student
$\pi_{\theta_0}$, frozen teacher $\pi_T$, curriculum parameters
$\tau_T,\tau_A,\tau_F,N_A,N_F$, competence criterion $\tau_S$,
and scheduled-step budget $K$

\Ensure Trained student $\pi_{\theta_K}$

\State Evaluate $q_T(x)$ and $q_{S,0}(x)$ for each
$x\in\mathcal{P}$

\State Construct $\mathcal{A}$ and $\mathcal{F}$ using
Equation~\ref{eq:anchor_frontier_space};
set $\mathcal{D}_{\mathrm{AF}}\gets\mathcal{A}\cup\mathcal{F}$

\State Construct and fix the schedule
$\mathcal{S}=(\mathcal{B}_0,\ldots,\mathcal{B}_{K-1})$
over $\mathcal{D}_{\mathrm{AF}}$ and the normalization denominators
$\{n_t\}_{t=0}^{K-1}$

\For{$t=0,\ldots,K-1$}

    \State Obtain the scheduled batch $\mathcal{B}_t$ from
    $\mathcal{S}$

    \State Sample
    $y_{t,i}\sim\pi_{\theta_t}(\cdot\mid x_i)$
    for each $x_i\in\mathcal{B}_t$

    \State Decode each trajectory and compute
    $q_{S,t}(x_i;y_{t,i})$ using
    Equation~\ref{eq:current_student_assessment}

    \State Form the effective batch
    \[
    \mathcal{B}^{\mathrm{eff}}_t
    \gets
    \left\{
    (x_i,y_{t,i})
    :
    x_i\in\mathcal{B}_t,\;
    q_{S,t}(x_i;y_{t,i})<\tau_S
    \right\}
    \]

    \If{$\mathcal{B}^{\mathrm{eff}}_t\neq\varnothing$}

        \State Score only the retained student trajectories
        with $\pi_T$

        \State Form $\mathcal{L}^{\mathrm{SCC}}_t(\theta)$ using
        Equation~\ref{eq:coupled_opd_objective}, with denominator
        $n_t$ and fixed trajectories and routing decisions

        \State
        $\theta_{t+1}
        \gets
        \mathcal{U}_{\mathrm{SCC}}
        \!\left(
        \theta_t,\mathcal{B}_t,
        \mathcal{B}^{\mathrm{eff}}_t;\pi_T
        \right)$

    \Else

        \State Skip optimizer and weight-decay updates;
        retain the optimizer state

        \State $\theta_{t+1}\gets\theta_t$

    \EndIf

\EndFor

\State \Return $\pi_{\theta_K}$

\end{algorithmic}
\end{algorithm}
\vspace{-0.8em}

\vspace{-0.4em}
\subsection{Curriculum Construction and Schedule Details}
\label{app:curriculum_details}

\vspace{-0.5em}
\paragraph{Curriculum construction.}
We construct AF offline from TimeLens-100K
\citep{zhang2025timelens}, using examples from
HiREST~\citep{zala2023hirest},
QuerYD~\citep{oncescu2021queryd},
CosMo-Cap~\citep{wang2024cosmo},
InternVid-VTime~\citep{wang2024internvid,huang2024vtimellm},
and DiDeMo~\citep{hendricks2017didemo}.
Following the construction in
Equation~\ref{eq:anchor_frontier_space}, we retain candidates with
$q_T(x)\geq0.7$ and select 100 Anchors with $q_{S,0}(x)\geq0.7$ and
900 Frontiers with $q_{S,0}(x)<0.3$.
Candidates in the intermediate competence range
$0.3\leq q_{S,0}(x)<0.7$ are excluded.
We use $0.7$ as a stringent task-level threshold for both teacher
qualification and Anchor membership, so that qualified candidates have
reliable teacher predictions and Anchors represent clearly supported
initial capabilities.
The Frontier threshold of $0.3$ isolates examples with substantial
learning headroom, while excluding the intermediate range creates a
clear separation between stabilization and acquisition.
Table~\ref{tab:af1k_sources} reports the resulting source composition.
\vspace{-1.5em}

\begin{table}[ht]
\centering
\scriptsize
\setlength{\tabcolsep}{2.7pt}
\caption{
\textbf{Source composition of AF.}
Unique counts refer to selected examples; presentations additionally
include four Frontier repeats in the final scheduled step.
}
\label{tab:af1k_sources}
\begin{tabular}{lrrrr}
\toprule
\textbf{Source}
& \textbf{Anchor}
& \textbf{Frontier}
& \textbf{Unique}
& \textbf{Presentations} \\
\midrule
CosMo-Cap       & 49 & 437 & 486 & 489 \\
InternVid-VTime & 26 & 235 & 261 & 262 \\
QuerYD         & 11 & 106 & 117 & 117 \\
DiDeMo         & 10 &  89 &  99 &  99 \\
HiREST         &  4 &  33 &  37 &  37 \\
\midrule
\textbf{Total}
& \textbf{100}
& \textbf{900}
& \textbf{1,000}
& \textbf{1,004} \\
\bottomrule
\end{tabular}
\vspace{-0.8em}
\end{table}


\vspace{-0.5em}
\paragraph{Training schedule.}
Per-step AF allocations follow a piecewise-linear profile with target shares of 100\%, 70\%, 30\%, 10\%, and 5\% at steps 1, 20, 40, 60, and 79, respectively. The interpolated allocations are normalized to 1,000 unique examples and deterministically rounded to integer counts, with a target Anchor--Frontier ratio of 1:9. Example ordering and assignment within these allocations are randomized before training and fixed throughout optimization. The final step includes four repeated Frontier presentations, yielding 1,004 presentations. For loss normalization, SCC uses $n_t=32$ for steps 1--78 and $n_t=8$ for step 79, independently of the number of assigned or retained AF presentations. Student-dependent realization selects trajectories for OPD without changing their assigned steps; steps with an empty effective batch advance the schedule without an optimizer update. Table~\ref{tab:af1k_schedule_distribution} summarizes presentation counts within each evaluation-checkpoint interval.



\vspace{-1.0em}
\begin{table}[ht]
\centering
\scriptsize
\setlength{\tabcolsep}{2.7pt}
\caption{
\textbf{Distribution of AF across the training schedule.}
Counts denote presentations grouped by evaluation-checkpoint intervals.
\textsuperscript{$\dagger$} includes four Frontier repeats in the final
step.
}
\label{tab:af1k_schedule_distribution}
\begin{tabular}{crrr}
\toprule
\textbf{Training Steps}
& \textbf{Anchor}
& \textbf{Frontier}
& \textbf{Total} \\
\midrule
1--20  & 53 & 481 & 534 \\
21--40 & 31 & 276 & 307 \\
41--59 & 12 & 104 & 116 \\
60--79 & 4  & $43^{\dagger}$ & 47 \\
\midrule
\textbf{Total}
& \textbf{100}
& $\mathbf{904}^{\dagger}$
& \textbf{1,004} \\
\bottomrule
\end{tabular}
\vspace{-1.2em}
\end{table}

\vspace{-0.4em}
\subsection{Detailed Accuracy--Efficiency Analysis}
\label{app:detailed_efficiency}
\vspace{-0.5em}

Table~\ref{tab:detailed_efficiency} complements
Figure~\ref{fig:training_efficiency} with training time and endpoint
recall.
We compare SCC with Video-OPD on its Teacher-Validated Disagreement
Focusing (TVDF) curriculum~\citep{li2026video} and on the same AF
curriculum.
Training time denotes wall-clock time on eight NVIDIA A100 GPUs,
including initialization, optimization, and checkpoint saving.
Mean recall at each threshold averages the corresponding metric across
Charades-TimeLens~\citep{zhang2025timelens}, ActivityNet-TimeLens~\citep{zhang2025timelens}, and
QVHighlights-TimeLens~\citep{zhang2025timelens}.

SCC reduces training time by
50.4\% relative to Video-OPD on TVDF and by 15.3\% on the same
AF curriculum, while improving mean recall at all three thresholds.
The AF comparison shows that the efficiency gain persists when the
candidate curriculum is held fixed.

AF construction uses teacher and initial-student predictions
scored against ground-truth intervals.
The resulting tIoU scores determine teacher qualification and the
Anchor and Frontier candidate regions, from which examples are selected
under the prescribed budgets.
Once these scores are available, selection requires no further model
inference, teacher scoring of student-generated prefixes, or token-level
divergence estimation.
\vspace{-1.2em}

\begin{table*}[ht]
\centering
\scriptsize
\setlength{\tabcolsep}{2.7pt}
\caption{
\textbf{Training time and endpoint recall.}
Video-OPD (TVDF) results are from \citet{li2026video}; other results are obtained locally. Training time on eight NVIDIA A100 GPUs is reported as hh:mm:ss. Recall (\%) is averaged across three TimeLens benchmarks before rounding to two decimals. Bold marks the shortest time and highest recall at each threshold.
}
\label{tab:detailed_efficiency}

\begin{tabular*}{\textwidth}{
    @{\extracolsep{\fill}}
    lrrrr
    @{}
}
\toprule
\textbf{Method}
& \textbf{Training Time}
& \textbf{Mean R@0.3}
& \textbf{Mean R@0.5}
& \textbf{Mean R@0.7} \\
\midrule

Video-OPD~\citep{li2026video} (TVDF)
& 2:01:46
& 69.13
& 50.57
& 39.53 \\

Video-OPD (AF)
& 1:11:20
& 70.69
& 52.53
& 40.70 \\

\textbf{SCC (AF)}
& \textbf{1:00:26}
& \textbf{71.83}
& \textbf{53.72}
& \textbf{41.84} \\

\bottomrule
\end{tabular*}
\vspace{-0.8em}
\end{table*}

\vspace{-0.4em}
\subsection{Comparison with TimeLens}
\label{app:timelens_comparison}
\vspace{-0.5em}

Table~\ref{tab:timelens_comparison} compares SCC with TimeLens-8B~\citep{zhang2025timelens} and single-round Video-OPD~\citep{li2026video}. TimeLens-8B undergoes TVG supervised fine-tuning followed by GRPO on 12,000 examples, while Video-OPD and SCC use OPD curricula of 2,500 and 1,000 examples, respectively. SCC outperforms single-round Video-OPD across all reported metrics. Compared with TimeLens-8B, SCC achieves higher scores on all ActivityNet-TimeLens metrics and higher mIoU, R@0.5, and R@0.7 on QVHighlights-TimeLens, including an R@0.7 of 54.3 versus 51.8. TimeLens-8B retains an advantage on Charades-TimeLens and QVHighlights-TimeLens R@0.3, indicating benchmark-dependent trade-offs across these training regimes.

\vspace{-1.2em}
\begin{table}[ht]
\centering
\caption{
\textbf{Comparison with TimeLens-8B.}
Baseline results are taken from Video-OPD~\citep{li2026video};
values are percentages;
bold indicates the best result in each column.
}
\label{tab:timelens_comparison}
\scriptsize
\setlength{\tabcolsep}{2.7pt}
\renewcommand{\arraystretch}{1.1}
\begin{tabular}{l|cccc|cccc|cccc}
\toprule
& \multicolumn{4}{c|}{Charades-TimeLens}
& \multicolumn{4}{c|}{ActivityNet-TimeLens}
& \multicolumn{4}{c}{QVHighlights-TimeLens} \\
\cmidrule(lr){2-5}
\cmidrule(lr){6-9}
\cmidrule(lr){10-13}
Method
& mIoU & R@0.3 & R@0.5 & R@0.7
& mIoU & R@0.3 & R@0.5 & R@0.7
& mIoU & R@0.3 & R@0.5 & R@0.7 \\
\midrule
TimeLens-8B
& \textbf{53.3} & \textbf{74.6} & \textbf{49.5} & \textbf{33.4}
& 49.3 & 63.8 & 48.0 & 36.4
& 63.0 & \textbf{77.8} & 63.4 & 51.8 \\
Video-OPD (Round 1)
& 52.0 & 73.1 & 45.8 & 32.4
& 47.3 & 60.5 & 45.6 & 35.8
& 61.0 & 73.8 & 60.3 & 50.4 \\
\rowcolor{gray!12}
SCC
& 52.8 & 74.0 & 48.1 & 33.0
& \textbf{50.6} & \textbf{64.8} & \textbf{49.2} & \textbf{38.2}
& \textbf{63.7} & 76.8 & \textbf{63.9} & \textbf{54.3} \\
\bottomrule
\end{tabular}
\vspace{-0.8em}
\end{table}

\subsection{Alternative Post-Training Objectives on AF}
\label{app:other_methods}
\vspace{-0.5em}

Table~\ref{tab:alternative_objectives} compares SCC with alternative post-training objectives~\citep{li2026video} on the same AF curriculum. SCC achieves the strongest R@0.7 on all three benchmarks and the best result across all Charades-TimeLens metrics. GRPO performs best on R@0.3 and R@0.5 for QVHighlights-TimeLens and on R@0.5 for ActivityNet-TimeLens, indicating different trade-offs between coarse retrieval and precise temporal localization. This comparison focuses on endpoint quality under method-specific training schedules. OP-FKD, OP-RKD, and GRPO are trained for 32 steps. GRPO runs for one epoch with a global prompt batch size of 32, sampling eight rollouts per example at temperature 1.0.


\vspace{-1.0em}

\begin{table*}[ht]
\centering
\scriptsize
\setlength{\tabcolsep}{6pt}

\caption{
\textbf{Alternative post-training objectives on AF.}
All methods use the AF curriculum and are trained and evaluated using
our local pipeline.
Off-policy forward- and reverse-KL distillation (OP-FKD and OP-RKD),
together with GRPO, follow the objective definitions in
Video-OPD~\citep{li2026video}.
All values are percentages, and bold indicates the best result in each
column.
}
\label{tab:alternative_objectives}

\resizebox{\textwidth}{!}{
\begin{tabular}{l|rrr|rrr|rrr}
\toprule
& \multicolumn{3}{c|}{\textbf{Charades-TimeLens}}
& \multicolumn{3}{c|}{\textbf{ActivityNet-TimeLens}}
& \multicolumn{3}{c}{\textbf{QVHighlights-TimeLens}} \\
\cmidrule(lr){2-4}
\cmidrule(lr){5-7}
\cmidrule(lr){8-10}

\textbf{Method}
& \textbf{R@0.3}
& \textbf{R@0.5}
& \textbf{R@0.7}
& \textbf{R@0.3}
& \textbf{R@0.5}
& \textbf{R@0.7}
& \textbf{R@0.3}
& \textbf{R@0.5}
& \textbf{R@0.7} \\
\midrule

OP-FKD~\citep{li2026video}
& 72.20
& 47.43
& 31.28
& 62.22
& 46.71
& 35.76
& 75.47
& 62.36
& 50.49 \\

OP-RKD~\citep{li2026video}
& 69.61
& 47.01
& 29.91
& 61.47
& 46.67
& 35.73
& 74.95
& 62.04
& 51.14 \\

GRPO~\citep{li2026video}
& 67.98
& 47.81
& 27.27
& 63.76
& \textbf{49.78}
& 36.31
& \textbf{77.16}
& \textbf{65.61}
& 51.53 \\

\textbf{SCC}
& \textbf{73.95}
& \textbf{48.14}
& \textbf{33.01}
& \textbf{64.78}
& 49.18
& \textbf{38.24}
& 76.77
& 63.85
& \textbf{54.25} \\

\bottomrule
\end{tabular}
}
\vspace{-1.2em}
\end{table*}

\subsection{Student-Dependent Realization on TVDF}
\label{app:tvdf_student_realization}
\vspace{-0.5em}
To assess student-dependent realization beyond the AF candidate space,
we retain the original TVDF curriculum and enable only the
current-student competence gate with $\tau_S=0.7$.
Of 2,504 scheduled presentations, 955 receive teacher scoring and OPD
updates, and 1,549 are suspended.
Table~\ref{tab:tvdf_student_realization} compares the variant at
checkpoint 79 with the published Video-OPD
results~\citep{li2026video}.

The variant achieves higher R@0.7 on all three benchmarks and improves
all three recall metrics on ActivityNet-TimeLens.
Small decreases occur at R@0.3 on Charades-TimeLens and at R@0.3 and
R@0.5 on QVHighlights-TimeLens.
These results support the applicability of student-dependent realization
beyond AF, with the most consistent gains under the stricter
localization criterion.

\vspace{-1.0em}

\begin{table*}[ht]
\centering
\scriptsize
\setlength{\tabcolsep}{5pt}
\renewcommand{\arraystretch}{1.08}

\caption{
\textbf{Student-dependent realization on TVDF.}
Both methods use the TVDF curriculum.
Video-OPD results are taken from \citet{li2026video}; the
student-dependent variant is trained and evaluated locally at
checkpoint 79.
Values are recall percentages rounded to one decimal place;
bold indicates the higher value in each column.
}
\label{tab:tvdf_student_realization}

\begin{tabularx}{\textwidth}{
    @{}
    >{\raggedright\arraybackslash}X
    |rrr|rrr|rrr
    @{}
}
\toprule
&
\multicolumn{3}{c|}{\textbf{Charades-TimeLens}}
&
\multicolumn{3}{c|}{\textbf{ActivityNet-TimeLens}}
&
\multicolumn{3}{c}{\textbf{QVHighlights-TimeLens}}
\\
\cmidrule(lr){2-4}
\cmidrule(lr){5-7}
\cmidrule(lr){8-10}

\textbf{Method}
& \textbf{R@0.3}
& \textbf{R@0.5}
& \textbf{R@0.7}
& \textbf{R@0.3}
& \textbf{R@0.5}
& \textbf{R@0.7}
& \textbf{R@0.3}
& \textbf{R@0.5}
& \textbf{R@0.7}
\\
\midrule

Video-OPD~\citep{li2026video}
& \textbf{73.1}
& 45.8
& 32.4
& 60.5
& 45.6
& 35.8
& \textbf{73.8}
& \textbf{60.3}
& 50.4
\\

Video-OPD + Student-Dep. Realization
& 72.7
& \textbf{46.2}
& \textbf{32.9}
& \textbf{62.1}
& \textbf{47.4}
& \textbf{37.4}
& 73.6
& 60.2
& \textbf{50.9}
\\

\bottomrule
\end{tabularx}
\vspace{-1.2em}
\end{table*}

\vspace{-0.4em}
\subsection{Sensitivity to the Student-Competence Criterion}
\label{app:student_criterion}
\vspace{-0.5em}

We vary the student-competence criterion over $\tau_S\in\{0.5,0.6,0.7,0.8,0.9\}$ while keeping the AF curriculum, student initialization, teacher, data order, and optimization protocol fixed. The main setting, $\tau_S = 0.7$, was fixed before benchmark evaluation and was not selected from this sweep. We set the main criterion to $\tau_S=0.7$ to apply the same stringent task-level standard when determining whether supervision remains necessary during training. All runs are evaluated at checkpoint 79, and a scheduled occurrence receives OPD supervision when $q_{S,t}<\tau_S$. Table~\ref{tab:threshold_routing} reports the resulting supervision routing, while Table~\ref{tab:threshold_performance} reports complete TVG performance.

\vspace{-1.2em}
\begin{table}[ht]
\centering
\scriptsize
\setlength{\tabcolsep}{5pt}
\caption{
\textbf{Realized supervision routing under different student-competence
criteria.}
All settings use the same 1,004 scheduled presentations.
The shaded row denotes the main setting.
}
\label{tab:threshold_routing}

\begin{tabular}{crrr}
\toprule
$\tau_S$
& \textbf{OPD Routes}
& \textbf{Suspended Routes}
& \textbf{OPD Route Rate} \\
\midrule

0.5
& 397
& 607
& 39.54\% \\

0.6
& 438
& 566
& 43.63\% \\

\rowcolor{gray!10}
\textbf{0.7}
& \textbf{492}
& \textbf{512}
& \textbf{49.00\%} \\

0.8
& 563
& 441
& 56.08\% \\

0.9
& 646
& 358
& 64.34\% \\

\bottomrule
\end{tabular}
\vspace{-2.0em}
\end{table}

\begin{table*}[ht]
\centering
\scriptsize
\setlength{\tabcolsep}{5pt}
\caption{
\textbf{TVG performance under different
student-competence criteria.}
All benchmarks use the TimeLens annotations. All settings use AF; ``--'' denotes uniform OPD supervision, while the remaining rows vary $\tau_S$. Values are percentages.
The shaded row denotes the setting used in the main experiments, and bold
indicates the best result in each column.
}
\label{tab:threshold_performance}

\resizebox{\textwidth}{!}{
\begin{tabular}{c|rrrr|rrrr|rrrr}
\toprule
& \multicolumn{4}{c|}{\textbf{Charades-TimeLens}}
& \multicolumn{4}{c|}{\textbf{ActivityNet-TimeLens}}
& \multicolumn{4}{c}{\textbf{QVHighlights-TimeLens}} \\
\cmidrule(lr){2-5}
\cmidrule(lr){6-9}
\cmidrule(lr){10-13}

\textbf{$\tau_S$}
& \textbf{mIoU}
& \textbf{R@0.3}
& \textbf{R@0.5}
& \textbf{R@0.7}
& \textbf{mIoU}
& \textbf{R@0.3}
& \textbf{R@0.5}
& \textbf{R@0.7}
& \textbf{mIoU}
& \textbf{R@0.3}
& \textbf{R@0.5}
& \textbf{R@0.7} \\
\midrule

-
& 52.26 
& 73.33 
& 47.13 
& 32.80
& 49.61 
& 63.40 
& 47.84 
& 37.20
& 62.49 
& 75.34 
& 62.62 
& 52.11 \\

0.5
& 52.51
& 73.71
& 47.81
& 32.68
& 50.52
& 64.20
& 48.82
& 38.22
& 63.17
& \textbf{76.83}
& 62.23
& 52.56 \\

0.6
& 52.41
& 73.68
& 47.07
& 32.47
& 50.30
& 63.84
& 48.36
& 38.18
& 63.30
& 76.18
& 62.69
& 53.41 \\

\rowcolor{gray!10}
\textbf{0.7}
& 52.79
& 73.95
& \textbf{48.14}
& 33.01
& \textbf{50.63}
& \textbf{64.78}
& \textbf{49.18}
& \textbf{38.24}
& \textbf{63.70}
& 76.77
& \textbf{63.85}
& \textbf{54.25} \\

0.8
& \textbf{52.85}
& \textbf{74.13}
& 47.79
& \textbf{33.51}
& 50.11
& 63.44
& 48.20
& 38.07
& 62.85
& 75.86
& 62.88
& 52.56 \\

0.9
& 52.40
& 73.39
& 47.81
& 32.35
& 49.67
& 63.20
& 47.62
& 37.38
& 61.93
& 74.56
& 62.10
& 51.66 \\

\bottomrule
\end{tabular}
}
\vspace{-1.2em}
\end{table*}

Increasing $\tau_S$ monotonically increases the fraction of teacher-scored OPD routes, from $39.54\%$ at $\tau_S=0.5$ to $64.34\%$ at $\tau_S=0.9$, whereas endpoint performance remains non-monotonic. The main setting, $\tau_S=0.7$, achieves the best result on eight of the twelve reported metrics, including all four ActivityNet-TimeLens metrics and three of the four QVHighlights-TimeLens metrics. Charades-TimeLens slightly favors $\tau_S=0.8$ on mean intersection over union (mIoU), R@0.3, and R@0.7. Performance declines at $\tau_S=0.9$ despite its higher supervision coverage, indicating that routing more examples through OPD does not necessarily improve the final student.

\vspace{-0.4em}
\subsection{Robustness to Teacher Choice}
\label{app:different_teachers}
\vspace{-0.5em}


We assess robustness to teacher choice using Qwen3-VL-4B and Qwen3-VL-8B~\citep{bai2025qwen3vl} teachers trained with the same GRPO recipe on a separately constructed 2,500-example difficulty-weighted curriculum. Construction follows the Time-R1 and TimeLens procedures~\citep{wang2025timer1,zhang2025timelens} adopted in Video-OPD~\citep{li2026video}, but the selected example set differs from that used to train its released 32B teacher. Each teacher supervises a Qwen3-VL-8B student on the original AF candidate set without teacher-specific re-filtering, with all other student-training settings held constant. We reuse the 32B-qualified AF set without assuming that every example remains qualified under the 4B/8B teachers. Table~\ref{tab:different_teachers} reports teacher and student performance across the three TVG benchmarks.

At the final scheduled step, students outperform their respective teachers in mIoU across all six teacher--benchmark pairings, supporting robustness to teacher choice. The 4B-supervised student leads on QVHighlights and ActivityNet, and the 8B-supervised student on Charades-STA, with no uniform benefit from greater teacher size. Differences in training procedures and, for the 4B teacher, model size preclude strictly matched before--after comparisons.

\vspace{-1em}
\begin{table*}[ht]
\centering
\scriptsize
\setlength{\tabcolsep}{5pt}
\renewcommand{\arraystretch}{1.05}
\caption{
\textbf{Robustness to teacher choice.}
GRPO teachers and their Qwen3-VL-8B students are evaluated at final
checkpoints using TimeLens annotations.
Student runs share the same AF configuration and differ only in teacher choice.
mIoU and recall are percentages; $\Delta$ mIoU denotes the student's
gain over its teacher in percentage points.
Bold marks the higher value for each metric within each teacher--student pair.
}
\label{tab:different_teachers}

\begin{tabular*}{\textwidth}{
@{\extracolsep{\fill}}
lllrrrrr
@{}
}
\toprule
\textbf{Teacher}
& \textbf{Model}
& \textbf{Benchmark}
& \textbf{mIoU}
& \textbf{R@0.3}
& \textbf{R@0.5}
& \textbf{R@0.7}
& \textbf{$\Delta$ mIoU} \\
\midrule

4B GRPO
& Direct teacher
& QVHighlights
& 61.61
& 76.51
& 61.65
& 50.49
& -- \\
4B GRPO
& \textbf{SCC student}
& QVHighlights
& \textbf{64.92}
& \textbf{77.87}
& \textbf{65.41}
& \textbf{55.09}
& \textbf{$+3.31$} \\

\addlinespace[1pt]

4B GRPO
& Direct teacher
& Charades-STA
& 51.35
& \textbf{73.15}
& 47.40
& 30.90
& -- \\
4B GRPO
& \textbf{SCC student}
& Charades-STA
& \textbf{52.39}
& 72.58
& \textbf{47.90}
& \textbf{33.66}
& \textbf{$+1.04$} \\

\addlinespace[1pt]

4B GRPO
& Direct teacher
& ActivityNet
& 49.83
& 64.82
& 48.49
& 37.02
& -- \\
4B GRPO
& \textbf{SCC student}
& ActivityNet
& \textbf{51.45}
& \textbf{65.29}
& \textbf{49.91}
& \textbf{39.20}
& \textbf{$+1.62$} \\

\midrule

8B GRPO
& Direct teacher
& QVHighlights
& 60.55
& 74.24
& 60.68
& 49.84
& -- \\
8B GRPO
& \textbf{SCC student}
& QVHighlights
& \textbf{62.98}
& \textbf{75.86}
& \textbf{63.01}
& \textbf{53.54}
& \textbf{$+2.43$} \\

\addlinespace[1pt]

8B GRPO
& Direct teacher
& Charades-STA
& 52.20
& 73.65
& 48.32
& 31.64
& -- \\
8B GRPO
& \textbf{SCC student}
& Charades-STA
& \textbf{53.05}
& \textbf{73.86}
& \textbf{48.68}
& \textbf{33.81}
& \textbf{$+0.85$} \\

\addlinespace[1pt]

8B GRPO
& Direct teacher
& ActivityNet
& 50.32
& 64.73
& 49.71
& 37.44
& -- \\
8B GRPO
& \textbf{SCC student}
& ActivityNet
& \textbf{51.11}
& \textbf{65.11}
& \textbf{49.73}
& \textbf{38.84}
& \textbf{$+0.79$} \\

\bottomrule
\end{tabular*}
\vspace{-0.8em}
\end{table*}

\vspace{-0.4em}
\subsection{Checkpoint-Wise TVG Performance}
\label{app:checkpoint_performance}
\vspace{-0.5em}

Table~\ref{tab:checkpoint_performance} summarizes TVG performance
across training checkpoints.
The final checkpoint achieves the highest mIoU on Charades-TimeLens
and QVHighlights-TimeLens, along with the highest R@0.7 on
QVHighlights-TimeLens.
ActivityNet-TimeLens reaches its best scores at checkpoints 40 and 59,
with modest declines at the final checkpoint.

\vspace{-1em}
\begin{table*}[ht]
\centering
\scriptsize
\setlength{\tabcolsep}{5pt}
\caption{
\textbf{Checkpoint-wise TVG performance of SCC.}
All checkpoints are taken from the main training run.
All values are percentages, and bold indicates the best checkpoint for
each metric within each benchmark.
}
\label{tab:checkpoint_performance}

\resizebox{\textwidth}{!}{
\begin{tabular}{c|rrrr|rrrr|rrrr}
\toprule
& \multicolumn{4}{c|}{\textbf{Charades-TimeLens}}
& \multicolumn{4}{c|}{\textbf{ActivityNet-TimeLens}}
& \multicolumn{4}{c}{\textbf{QVHighlights-TimeLens}} \\
\cmidrule(lr){2-5}
\cmidrule(lr){6-9}
\cmidrule(lr){10-13}

\textbf{Checkpoint}
& \textbf{mIoU}
& \textbf{R@0.3}
& \textbf{R@0.5}
& \textbf{R@0.7}
& \textbf{mIoU}
& \textbf{R@0.3}
& \textbf{R@0.5}
& \textbf{R@0.7}
& \textbf{mIoU}
& \textbf{R@0.3}
& \textbf{R@0.5}
& \textbf{R@0.7} \\
\midrule

20
& 48.24
& 68.66
& 45.97
& 28.84
& 49.39
& 63.93
& 49.07
& 37.31
& 62.80
& \textbf{77.35}
& \textbf{64.50}
& 51.59 \\

40
& 52.75
& \textbf{74.04}
& 47.78
& 32.92
& \textbf{50.93}
& 64.67
& \textbf{49.47}
& \textbf{38.64}
& 63.46
& 76.77
& 63.27
& 53.28 \\

59
& 52.73
& 73.60
& 47.81
& \textbf{33.24}
& 50.86
& \textbf{65.16}
& 49.31
& \textbf{38.64}
& 63.39
& 76.31
& 63.40
& 53.28 \\

79
& \textbf{52.79}
& 73.95
& \textbf{48.14}
& 33.01
& 50.63
& 64.78
& 49.18
& 38.24
& \textbf{63.70}
& 76.77
& 63.85
& \textbf{54.25} \\

\bottomrule
\end{tabular}
}
\vspace{-1.2em}
\end{table*}

\subsection{Sensitivity to the Anchor--Frontier Ratio}
\label{app:af_ratio}
\vspace{-0.5em}

We examine how the Anchor/Frontier ratio affects realized supervision
and endpoint performance within a fixed 1,000-example candidate space.
The main 1:9 configuration is compared with 3:7 and 1:1 alternatives. The main $1{:}9$ ratio was fixed before benchmark evaluation
and was not selected from this comparison.
Each configuration follows the same student-dependent realization rule
and is evaluated at checkpoint 79.
Table~\ref{tab:af_ratio} reports the supervision routing and
TVG performance.

The region-specific routing pattern remains stable across the tested
ratios.
Only 11.0--13.6\% of Anchor presentations receive OPD supervision,
compared with 53.2--53.9\% of Frontier presentations.
As the Anchor share increases, the overall OPD route rate decreases
from 49.0\% to 41.8\% and 33.8\%.
The candidate-space ratio therefore directly shapes the realized
supervision workload under the same routing rule.

The main 1:9 ratio achieves the best result on 11 of the 12 TVG
metrics.
Both alternative ratios use fewer teacher-scored OPD routes than the
main configuration and yield lower endpoint performance overall.
Together, these results show that the Anchor--Frontier ratio controls
an accuracy--workload trade-off, with the Frontier-heavy main
configuration providing the strongest endpoint performance among the
tested ratios.

\vspace{-1.2em}

\begin{table*}[ht]
\centering
\scriptsize
\setlength{\tabcolsep}{5pt}
\renewcommand{\arraystretch}{1.05}

\caption{
\textbf{Sensitivity to the Anchor--Frontier ratio.}
Each candidate space contains 1,000 unique examples.
Ratios denote the numbers of unique Anchor and Frontier examples.
Panel (a) reports routing over 1,004 scheduled presentations, including
repeated occurrences in the training schedule.
Panel (b) reports TVG performance at checkpoint 79.
Values are percentages.
The shaded rows denote the main 1:9 configuration, and bold values in
panel (b) indicate the best result in each column.
}
\label{tab:af_ratio}
\label{tab:realized_supervision_routes}

\textit{(a) Realized supervision routing.}

\vspace{0.25em}

\begin{tabular*}{\textwidth}{
    @{\extracolsep{\fill}}
    l|rrr|rrr|rrr
    @{}
}
\toprule
&
\multicolumn{3}{c|}{\textbf{Anchor}}
&
\multicolumn{3}{c|}{\textbf{Frontier}}
&
\multicolumn{3}{c}{\textbf{Overall}}
\\
\cmidrule(lr){2-4}
\cmidrule(lr){5-7}
\cmidrule(lr){8-10}

\textbf{A/F ratio}
& \textbf{OPD Routes}
& \textbf{Suspended}
& \textbf{Rate}
& \textbf{OPD Routes}
& \textbf{Suspended}
& \textbf{Rate}
& \textbf{OPD Routes}
& \textbf{Suspended}
& \textbf{Rate}
\\
\midrule

\rowcolor{gray!10}
\textbf{1:9}
& 11
& 89
& 11.00\%
& 481
& 423
& 53.21\%
& 492
& 512
& 49.00\%
\\

3:7
& 41
& 260
& 13.62\%
& 379
& 324
& 53.91\%
& 420
& 584
& 41.83\%
\\

1:1
& 68
& 433
& 13.57\%
& 271
& 232
& 53.88\%
& 339
& 665
& 33.76\%
\\

\bottomrule
\end{tabular*}

\vspace{0.8em}

\textit{(b) Endpoint TVG performance.}

\vspace{0.25em}

\resizebox{\textwidth}{!}{
\begin{tabular}{l|rrrr|rrrr|rrrr}
\toprule
&
\multicolumn{4}{c|}{\textbf{Charades-TimeLens}}
&
\multicolumn{4}{c|}{\textbf{ActivityNet-TimeLens}}
&
\multicolumn{4}{c}{\textbf{QVHighlights-TimeLens}}
\\
\cmidrule(lr){2-5}
\cmidrule(lr){6-9}
\cmidrule(lr){10-13}

\textbf{A/F ratio}
& \textbf{mIoU}
& \textbf{R@0.3}
& \textbf{R@0.5}
& \textbf{R@0.7}
& \textbf{mIoU}
& \textbf{R@0.3}
& \textbf{R@0.5}
& \textbf{R@0.7}
& \textbf{mIoU}
& \textbf{R@0.3}
& \textbf{R@0.5}
& \textbf{R@0.7}
\\
\midrule

\rowcolor{gray!10}
\textbf{1:9}
& \textbf{52.79}
& \textbf{73.95}
& \textbf{48.14}
& \textbf{33.01}
& \textbf{50.63}
& \textbf{64.78}
& \textbf{49.18}
& 38.24
& \textbf{63.70}
& \textbf{76.77}
& \textbf{63.85}
& \textbf{54.25}
\\

3:7
& 52.35
& 73.63
& 47.04
& 32.77
& 49.95
& 63.49
& 47.98
& 38.04
& 62.32
& 74.69
& 62.04
& 52.50
\\

1:1
& 52.43
& 73.36
& 47.10
& 32.80
& 50.39
& 64.11
& 48.29
& \textbf{38.36}
& 62.86
& 75.86
& 62.43
& 52.95
\\

\bottomrule
\end{tabular}
}
\vspace{-1.0em}
\end{table*}

\vspace{-0.4em}
\subsection{Sensitivity to Candidate-Space Scale}
\label{app:af_scale}
\vspace{-0.5em}
We examine sensitivity to candidate-space scale by comparing the main 1,000-example AF curriculum with 1,500-, 2,000-, and 2,500-example variants, all maintaining an Anchor--Frontier ratio of $1{:}9$. All runs use 79 scheduled steps with global batch sizes capped at 32. Larger curricula preserve the original 1,000 examples at their scheduled positions and add teacher-qualified examples through a deterministic, source-matched allocation balanced across vacant positions. All unique examples are presented, with four final-step repeats yielding 1,004, 1,504, 2,004, and 2,504 presentations, respectively. Table~\ref{tab:af_scale} reports TVG performance at checkpoint 79.

Expanding the candidate space does not consistently improve endpoint performance. The main 1,000-example configuration achieves the best result on nine of the twelve metrics, including all four on QVHighlights-TimeLens. The 1,500-example variant yields small gains in Charades-TimeLens R@0.3 and ActivityNet-TimeLens mIoU and R@0.7. These results show that a compact AF curriculum can define an effective supervision space for OPD.

\vspace{-1.2em}

\begin{table*}[ht]
\centering
\scriptsize
\setlength{\tabcolsep}{5pt}
\renewcommand{\arraystretch}{1.05}

\caption{
\textbf{Sensitivity to candidate-space scale.}
Candidate-space size counts unique examples.
All configurations use an Anchor--Frontier ratio of $1{:}9$ and
are evaluated at checkpoint 79.
Values are percentages rounded to two decimal places.
The shaded row denotes the main configuration, and bold indicates
the best result in each column.
}
\label{tab:af_scale}

\resizebox{\textwidth}{!}{
\begin{tabular}{l|rrrr|rrrr|rrrr}
\toprule
&
\multicolumn{4}{c|}{\textbf{Charades-TimeLens}}
&
\multicolumn{4}{c|}{\textbf{ActivityNet-TimeLens}}
&
\multicolumn{4}{c}{\textbf{QVHighlights-TimeLens}}
\\
\cmidrule(lr){2-5}
\cmidrule(lr){6-9}
\cmidrule(lr){10-13}

\shortstack[l]{\textbf{Candidate-Space}}
& \textbf{mIoU}
& \textbf{R@0.3}
& \textbf{R@0.5}
& \textbf{R@0.7}
& \textbf{mIoU}
& \textbf{R@0.3}
& \textbf{R@0.5}
& \textbf{R@0.7}
& \textbf{mIoU}
& \textbf{R@0.3}
& \textbf{R@0.5}
& \textbf{R@0.7}
\\
\midrule

\rowcolor{gray!10}
\textbf{1,000}
& \textbf{52.79}
& 73.95
& \textbf{48.14}
& \textbf{33.01}
& 50.63
& \textbf{64.78}
& \textbf{49.18}
& 38.24
& \textbf{63.70}
& \textbf{76.77}
& \textbf{63.85}
& \textbf{54.25}
\\

1,500
& 52.69
& \textbf{74.19}
& 47.58
& 32.65
& \textbf{50.81}
& 64.40
& 49.07
& \textbf{38.56}
& 62.40
& 75.41
& 61.71
& 51.79
\\

2,000
& 52.63
& 74.07
& 47.61
& 32.83
& 50.07
& 63.82
& 48.13
& 37.91
& 62.70
& 75.67
& 62.43
& 53.08
\\

2,500
& 52.00
& 73.51
& 47.04
& 31.91
& 50.38
& 64.11
& 48.82
& 38.31
& 62.76
& 76.38
& 62.82
& 52.43
\\

\bottomrule
\end{tabular}
}
\vspace{-0.8em}
\end{table*}

\vspace{-0.4em}
\subsection{Additional Controls on Student-Dependent Realization}
\label{app:realization_controls}
\vspace{-0.5em}
To better understand SCC's gains, we evaluate a budget-matched static
routing baseline and a variant using effective-batch normalization.
Table~\ref{tab:realization_controls} reports results at checkpoint 79.

\vspace{-0.5em}

\paragraph{Budget-matched static routing.}
We fix a routing mask before training to assign teacher scoring and OPD to 492 of the 1,004 scheduled presentations, matching SCC's realized supervision budget. Constrained random sampling selects 49 distinct Anchors and 443 distinct Frontiers, preserving the $1{:}9$ Anchor--Frontier ratio and within-region source proportions up to integer rounding. Selected examples retain their original positions, with per-step supervision counts matched using only aggregate counts from the reference SCC run. The mask remains fixed throughout the control run, and loss normalization uses the same per-step denominators as SCC. SCC achieves higher mIoU on all three benchmarks and outperforms this control on 11 of the 12 metrics.

\vspace{-0.5em}
\paragraph{Effective-batch normalization.}
This experiment retains student-dependent routing and averages the OPD loss over retained presentations using the active-only normalization defined in Appendix~\ref{app:normalization_execution}. The main SCC configuration performs better on most metrics.

\vspace{-0.8em}
\begin{table*}[ht]
\centering
\scriptsize
\setlength{\tabcolsep}{5pt}
\renewcommand{\arraystretch}{1.05}

\caption{
\textbf{Controls on student-dependent realization.}
All runs use AF and are evaluated at checkpoint 79.
Static routing fixes 492 supervised presentations before training.
SCC (effective-batch) averages loss over supervised presentations per batch.
Values are percentages; bold marks column maxima and shading denotes SCC.
}
\label{tab:realization_controls}

\resizebox{\textwidth}{!}{
\begin{tabular}{l|rrrr|rrrr|rrrr}
\toprule
&
\multicolumn{4}{c|}{\textbf{Charades-TimeLens}}
&
\multicolumn{4}{c|}{\textbf{ActivityNet-TimeLens}}
&
\multicolumn{4}{c}{\textbf{QVHighlights-TimeLens}}
\\
\cmidrule(lr){2-5}
\cmidrule(lr){6-9}
\cmidrule(lr){10-13}

\textbf{Configuration}
& \textbf{mIoU}
& \textbf{R@0.3}
& \textbf{R@0.5}
& \textbf{R@0.7}
& \textbf{mIoU}
& \textbf{R@0.3}
& \textbf{R@0.5}
& \textbf{R@0.7}
& \textbf{mIoU}
& \textbf{R@0.3}
& \textbf{R@0.5}
& \textbf{R@0.7}
\\
\midrule

Video-OPD on AF
& 52.26
& 73.33
& 47.13
& 32.80
& 49.61
& 63.40
& 47.84
& 37.20
& 62.49
& 75.34
& 62.62
& 52.11
\\

\addlinespace[2pt]

\shortstack[l]{Static routing}
& 52.21
& 72.79
& 47.19
& \textbf{33.04}
& 49.65
& 63.36
& 47.29
& 37.44
& 62.12
& 74.95
& 61.39
& 51.53
\\

\addlinespace[2pt]

\shortstack[l]{SCC (effective-batch)}
& \textbf{52.82}
& \textbf{74.31}
& 48.02
& 32.74
& 50.55
& 64.53
& 48.78
& 38.13
& 62.88
& 75.54
& 62.75
& 53.54
\\

\addlinespace[2pt]

\rowcolor{gray!10}
\textbf{SCC}
& 52.79
& 73.95
& \textbf{48.14}
& 33.01
& \textbf{50.63}
& \textbf{64.78}
& \textbf{49.18}
& \textbf{38.24}
& \textbf{63.70}
& \textbf{76.77}
& \textbf{63.85}
& \textbf{54.25}
\\

\bottomrule
\end{tabular}
}
\vspace{-0.5em}
\end{table*}

\begin{figure*}[ht]
\centering
\includegraphics[width=\textwidth]{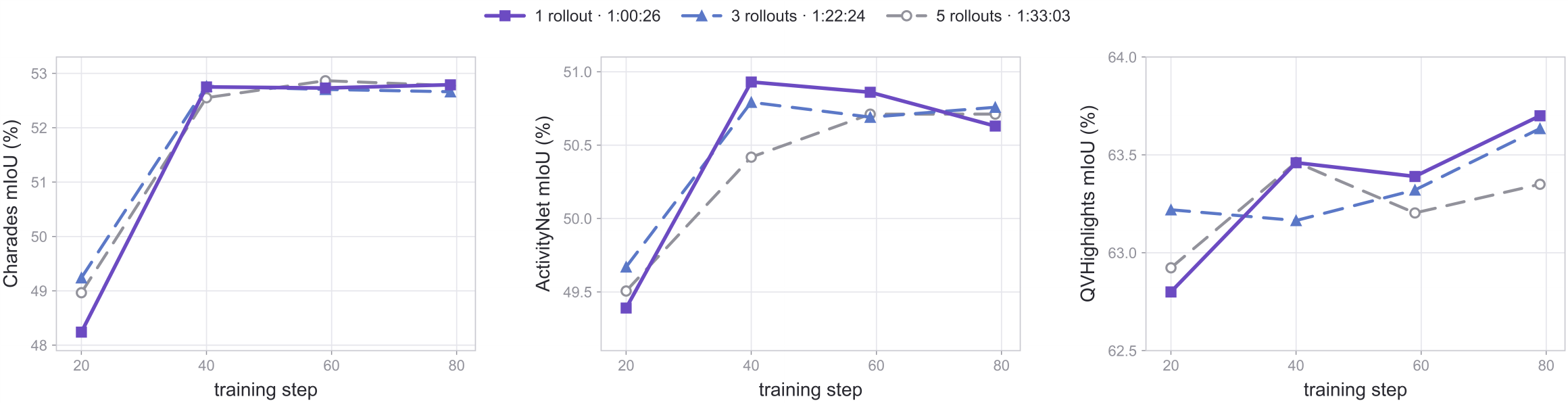}
\vspace{-1em}
\caption{
\textbf{Sensitivity to rollout count.}
Benchmark mIoU during training with one, three, or five student rollouts per scheduled example. Total training times are 1:00:26, 1:22:24, and 1:33:03, respectively (hours:minutes:seconds).}
\label{fig:rollout_count}
\vspace{-0.8em}
\end{figure*}

\vspace{-0.4em}
\subsection{Sensitivity to Rollout Count}
\label{app:rollout_count}
\vspace{-0.5em}

We compare the default single-rollout assessment with three- and five-rollout variants using mean tIoU against the ground-truth interval. When this score falls below $\tau_S=0.7$, only the first sampled trajectory receives teacher scoring and contributes to the OPD loss. Figure~\ref{fig:rollout_count} reports benchmark mIoU across training checkpoints.

Additional rollouts provide modest gains at some checkpoints, including slightly higher endpoint mIoU on ActivityNet-TimeLens, but yield no consistent advantage across benchmarks. Training time increases with the rollout count. These results support single-rollout assessment as a practical choice, maintaining comparable overall TVG performance with shorter training time under the evaluated settings.

\vspace{-0.4em}
\subsection{General Video Understanding}
\label{app:general_understanding}
\vspace{-0.5em}
We further evaluate broader video understanding on
TempCompass~\citep{liu2024tempcompass},
MVBench~\citep{li2024mvbench}, and
Video-MME~\citep{fu2025videomme}, using accuracy as the evaluation
metric.
All post-training runs and evaluations use our local pipeline.
The baselines follow the objective definitions in
Video-OPD~\citep{li2026video} and are trained on the same AF
curriculum.

As shown in Table~\ref{tab:general_video_results}, SCC achieves the
highest accuracy on all three benchmarks.
Although the gains are modest, their consistency indicates that the
improvements in temporal grounding do not compromise broader
video-understanding capabilities.

\vspace{-1em}

\begin{table}[ht]
    \centering
    \scriptsize
    \setlength{\tabcolsep}{5pt}
    \caption{
    \textbf{Evaluation on broader video-understanding benchmarks.}
    Values are accuracy percentages, and bold indicates the best result
    in each column.
    }
    \label{tab:general_video_results}
    \begin{tabular}{lccc}
        \toprule
        \textbf{Method}
        & \textbf{TempCompass}
        & \textbf{MVBench}
        & \textbf{Video-MME} \\
        \midrule

        Qwen3-VL-8B-Instruct~\citep{bai2025qwen3vl}
        & 73.16
        & 68.17
        & 67.96 \\

        OP-RKD~\citep{li2026video}
        & 73.16
        & 68.73
        & 67.52 \\

        OP-FKD~\citep{li2026video}
        & 73.29
        & 68.15
        & 67.33 \\

        GRPO~\citep{li2026video}
        & 73.04
        & 68.95
        & 67.63 \\

        \textbf{SCC}
        & \textbf{73.35}
        & \textbf{69.08}
        & \textbf{68.56} \\

        \bottomrule
    \end{tabular}
    \vspace{-0.8em}
\end{table}

\end{document}